%% file: main.tex
\documentclass{article}

\usepackage{misc/acl_2023}

\usepackage[utf8]{inputenc} 
\usepackage[T1]{fontenc}    
\usepackage{hyperref}       
\usepackage{url}            
\usepackage{booktabs}       
\usepackage{amsfonts}       
\usepackage{nicefrac}       
\usepackage{microtype}      

\usepackage[]{algorithm2e}
\usepackage{times}
\usepackage{latexsym}
\usepackage{graphicx}
\usepackage{amsmath,amssymb}
\usepackage{cleveref}
\usepackage{caption}
\usepackage{subcaption}
\usepackage{enumitem}
\usepackage{bm}
\usepackage{soul}
\usepackage{svg}
\usepackage{amsmath}
\usepackage{color, colortbl}
\definecolor{Gray}{gray}{0.9}
\definecolor{LightCyan}{rgb}{0.88,1,1}
\definecolor{LGreen}{rgb}{0.5,0.9,0.5}
\definecolor{MGreen}{rgb}{0.3,0.9,0.3}
\definecolor{LRed}{rgb}{0.9,0.5,0.5}
\definecolor{Pronoun}{rgb}{0.9,0.5,0.5}

\usepackage[normalem]{ulem} 
\usepackage{xcolor}

\usepackage[T1]{fontenc}


\usepackage[normalem]{ulem} 
\usepackage{xcolor}

\begin{document}

\title{Understanding the role of FFNs in driving multilingual behavior in LLMs}

%

\author{%
  Sunit Bhattacharya and Ond\v{r}ej Bojar\\
  Institute of Formal and Applied Linguistics\\
  Faculty of Mathematics and Physics\\
  Charles University\\
  \texttt{(bhattacharya,bojar)@ufal.mff.cuni.cz} \\
}

\maketitle

\begin{abstract}
Multilingualism in Large Language Models (LLMs) is an yet under-explored area. In this paper, we conduct an in-depth analysis of the multilingual capabilities of a family of a Large Language Model, examining its architecture, activation patterns, and processing mechanisms across languages. We introduce novel metrics to probe the model's multilingual behaviour at different layers and shed light on the impact of architectural choices on multilingual processing. 

Our findings reveal different patterns of multilinugal processing in the sublayers of Feed-Forward Networks of the models. Furthermore, we uncover the phenomenon of "over-layerization" in certain model configurations, where increasing layer depth without corresponding adjustments to other parameters may degrade model performance. Through comparisons within and across languages, we demonstrate the interplay between model architecture, layer depth, and multilingual processing capabilities of LLMs trained on multiple languages. 

\end{abstract}

\setlength{\parskip}{0pt}

\input{content.tex}

\newpage
\bibliography{misc/bibliography}
\bibliographystyle{misc/acl_natbib}

\section*{Appendix}

\appendix

\input{appendix.tex}
\end{document}

%% file: content.tex
\section{Introduction}
\label{intro}

Large Language Models (LLMs) are consistently getting better at multilingual NLP, e.g. at machine translation (MT) or performing cross-lingual tasks. But is this enough to conclude that these models are learning representations or patterns that generalise across multiple languages? Such an expectation is an old one, dating back to the proposal of ``universals of language'' \cite{greenberg1966language}. Previous work (see \Cref{related}) with word embeddings and intermediate representations of layers learnt by auto-encoding models such as BERT \cite{rogers2021primer} has shown some degree of alignment in internal representation among languages. However, similar analysis of decoder-only Transformer models and especially their Feed-Forward Network (FFN) layers is so far insufficient. 

In this work, we focus on the FFNs of large autoregressive language models. FFNs are characterised by their layer-by-layer processing of input representations and make up the bulk of the total parameters of all LLMs. They process individual tokens in parallel, with all cross-token information transfer happening in the self-attention layers.  Still, despite their seemingly straightforward structure, their function in Transformer models has not been fully understood yet. Although recent work has explored how the FFNs process layer-wise intermediate representations leading to the next word prediction in LLMs, there are still many unanswered questions. Multilingualism (how different languages are represented inside the language models) in FFNs is one such topic.

\begin{figure}[htbp]
    \centering
    \includegraphics[scale=0.9]{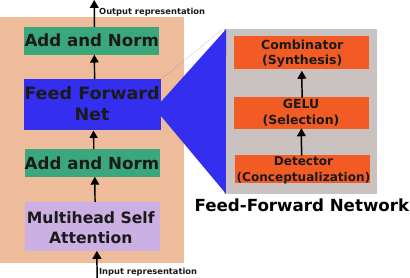}
    \caption{Transformer block and the structure of FFN}
    \label{fig:ffn-diagram}
\end{figure}

In this work, we build upon the view of the FFNs introduced by \citet{bhattacharya2023unveiling}. We consider the lower sublayer of the FFN as \emph{detectors} and the higher sublayer as \emph{combinators} (\Cref{fig:ffn-diagram}) (\Cref{fig:ffn-diagram}). Over the duration of training, the detectors at the different layers start getting triggered by specific patterns in the input data \citep{geva2021transformer}. The activation function ``selects'' the important aspects (selection) and the combinators ``combine'' them to emit an output which can be interpreted as a prediction of the planned next token for that layer \cite{belrose2023eliciting,geva2022transformer,dar2023analyzing}. In fact, the final prediction in the Transformer models might result from a process of ``incremental prediction'' (based on the idea of iterative inference in \citealp{jastrzkebski2017residual}) through the layers. Hence, during training, the model is trained to slowly convert the input representations close to the expected output, one layer (processing step) at a time. Thus, the intermediate activations during inference can reveal how input representations evolve within the model, mapping input prefixes to predicted output.

In this paper:
\begin{itemize}
    \item Refine the analysis method of \citet{bhattacharya2023unveiling} and propose a novel measure called activation flatness to identify and study activity patterns corresponding to multiple languages in LLMs.
    \item Describe how this activity varies across model layers and across model sizes.
    \item Identify distinct zones of multilingual and language-specific sub-layers in the FFNs. 
\end{itemize}
 
We study the working of detectors and combinators across layers of four XGLM models \cite{lin2022few} (568M, 1.7B, 2.9B, 7.5B) which are freely available, share the same architecture (decoder-only Transformer) and are trained on the same data comprising of 500B tokens from 30 diverse languages. We select this particular model family because the training data curation involved up-sampling of under-resourced languages to achieve a balanced language representation. This makes the models perfect candidates to study multilingual representations in LLMs.

\section{Related Work}
\label{related}
\paragraph{Multilingual alignment in LLMs.}
There is no consensus yet on multilinugal LLMs learning universal patterns across languages. However there is clear evidence that they learn embeddings which have high overlap across languages, primarily between those of the same family \cite{doddapaneni2021primer}.  \citet{muller2021first} show that a multilinugal BERT model can be seen as a stack of a language-specific encoder at the lower layers and a language-agnsotic predictor in the upper layers. The language agnosticity of the upper layers was also confirmed by \cite{liu2020study,pires2019multilingual}. \cite{stanczak2022same} have demonstrated significant cross-lingual overlap between neurons in multilingual auto-encoder language models. A probing based analysis of multilinguality of autoregressive models was done by \citet{mueller2022causal}. But on the question of language models learning universal patters, recently \citet{gurnee2024universal} have demonstrated the existance of universal neurons in GPT2 models.  \cite{xie2021importance} attempted to identify language-specific neurons in an encoder-decoder based machine translation systems using Transformers. 

\paragraph{Model Sparsity.}
In recent literature \cite{liu2023deja,mirzadeh2023relu}, sparsity has been thought of as a way to increase the efficiency of models during inference. To do that, analyses of models employing the ReLU activation functions have been done. Some works have also explored the ``ReLU-fication'' \cite{song2024prosparse} of pretrained networks. Sparse models also should be easier to interpret \cite{tamkin2023codebook}.

\section{Methodology}
\label{methoodlogy}
\subsection{Model snapshots}

In order to better understand how language models arrive at their prediction during inference, we collect ``model snapshots'', i.e. intermediate representations of the layers of different models of the XGLM family. To do that, we feed the language models with prefixes of sentences one word at a time and ask them to predict the next word. 

For an English sentence like ``Elementary, my dear Watson.'', we first remove all punctuations except the last i.e. period in this case. For this sentence, we obtain 8 subwords: [`Elementar', `y', `my', `de', `ar', `Watson', `.'].\footnote{Note that the subword model used in XGLM does not distinguish between word-internal and word-final subwords, increasing the conflation of tokens across languages (e.g. ``a'' being the determiner in English and a conjunction in Czech)} The prefixes (at the subword level) would thus be: [Elementar', `y'], [`Elementar', `y', `my'], [`Elementar', `y', `my', `de', `ar'], [`Elementar', `y', `my', `de', `ar', `Watson'] and [`Elementar', `y', `my', `de', `ar', `Watson', `.'].

\subsection{Snapshots from parallel test-sets}

Corresponding to each prefix, we save the model snapshots of all the layers, focusing only on the FFN layers (i.e. the detector and combinator sublayers) and discarding everything else. Since we are explicitly interested in the multilingual aspects of the model(s), we feed them with prefixes obtained from sentences across 4 different WMT\footnote{https://machinetranslate.org/wmt} test-sets. Each test set contains parallel sentences in English and another language (Czech, French, German or Hindi). Hence, for each language pair, we end up with model snapshots for prefixes of both languages.

Thus, each prefix yields a \emph{detector} and \emph{combinator} representation for each layer of the model in the form of a tensor of shape $(l,d)$ where $l$ indicates the number of subwords in the prefix and $d$ indicates the dimension of the layer. So, for a layer of the model with 1024 neurons, a prefix with 4 subwords would yield a vector of shape $(4,1024)$. This notation will be relevant later in the paper. For the most of our analyses, we extract the representation of the last subword of the prefix assuming that it contributes most closely to the prediction of the next word, leading us to a vector of shape (1,1024). Later, in \Cref{activation_flatness}, we will use of all the prefix representations.

\subsection{Focus of Interest}

Our primary objective is to reveal and understand ``concentrated activity'' across the layers in LLMs in the context of multilingual language processing. We do it in three levels of granularity. We start by observing the \textbf{patterns of sparsity} in the models across languages. Then we dive deeper and focus on the \textbf{nature of distribution} of the activations. And finally, we determine the degree of multilinguality of the neurons through the layers by looking at \textbf{patterns of shared activation}.

\section{Observations}
\label{observations}

\subsection{Sparsity patterns across languages}
We start with a very basic analysis of the activations of detectors and combinators of the model. The XGLM models utilize the GELU activation function which, similarly to ReLU, allows the training to reach a sparse representation. Thus, we start with examining if the processing of certain languages correspond to sparser representations. In order to assess the level of sparsity across model sizes and layers, we use activation frequency.

Given a set of model snapshots over a set of input prefixes, we define the activation frequency for a particular neuron as the ratio of activation counts (number of instances where the activation value was non-zero) to the total number of prefixes. This gives an idea about the overall importance of this neuron for the given test inputs. Activation frequency close to one indicates that the neuron has fired for almost all prefixes. For each layer, we collect activation frequencies of all the neurons and consider its average and, more importantly, standard deviation. Highly varying activation frequency across neurons of a layer indicates that some neurons are very important and some are very unimportant for the given set of prefixes, i.e. that the overall representation is sparse. 

\begin{figure}[ht]  
    \centering
    \begin{subfigure}[b]{0.2\textwidth}
        \centering
        \includegraphics[scale=0.28]{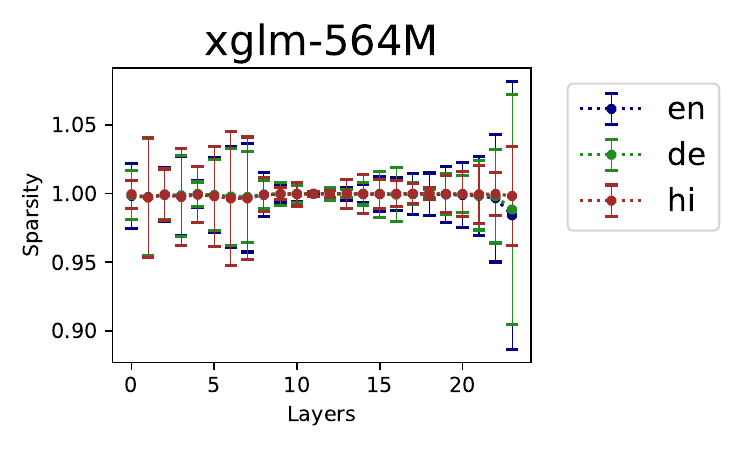}
    \end{subfigure}
    \hfill 
    \begin{subfigure}[b]{0.2\textwidth}
        \centering
        \includegraphics[scale=0.28]{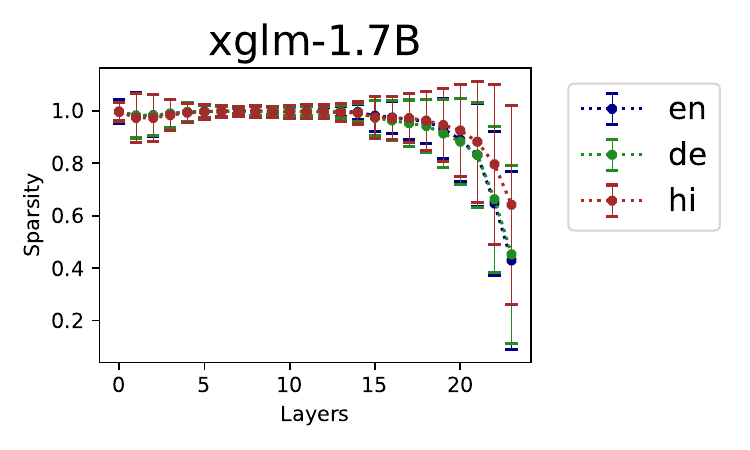}   
    \end{subfigure}
    \begin{subfigure}[b]{0.2\textwidth}
        \centering
        \includegraphics[scale=0.28]{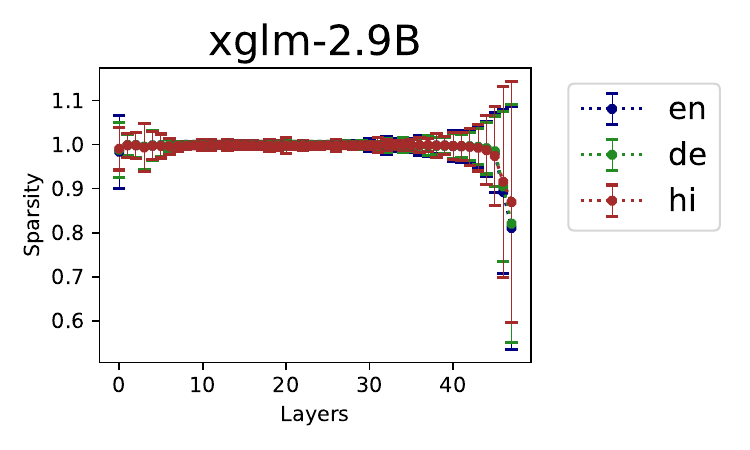}
    \end{subfigure}
    \hfill
    \begin{subfigure}[b]{0.2\textwidth}
        \centering
        \includegraphics[scale=0.28]{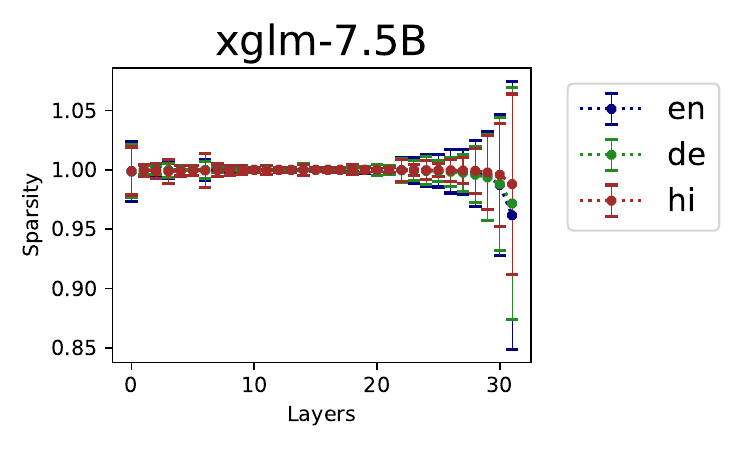}
    \end{subfigure}   
    \caption{Activation frequency for detectors along with standard deviation plotted for English, German and Hindi.}
    \label{fig:det_act_freq}
\end{figure}

We present the average and standard deviation of activation frequency of detectors across the studied model sizes in \Cref{fig:det_act_freq}

We observe that for the detectors, the average activation frequency drops near the output layers for all languages. But it remains stable (and close to 1) for all other layers. Also, the standard deviation of activation frequencies increase near the input and output layers. In other words, input and output layers have far sparser detector representations than the middle layers. We also observe that the representations in middle layers become increasingly dense (smaller standard deviations) with increase in model size. Finally, the greatest drop in the activation frequency occurs for languages like English, French and German. In \Cref{fig:det_act_freq}, we show results from only three languages (i.e. English, German and Hindi) for readability.

\begin{figure}[ht]  
    \centering 
    \begin{subfigure}[b]{0.22\textwidth}
        \centering
        \includegraphics[scale=0.28]{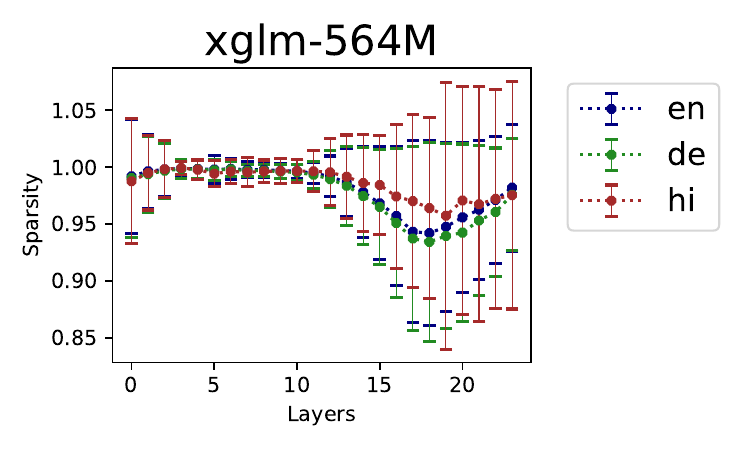}
    \end{subfigure}
    \hfill 
    \begin{subfigure}[b]{0.22\textwidth}
        \centering
        \includegraphics[scale=0.28]{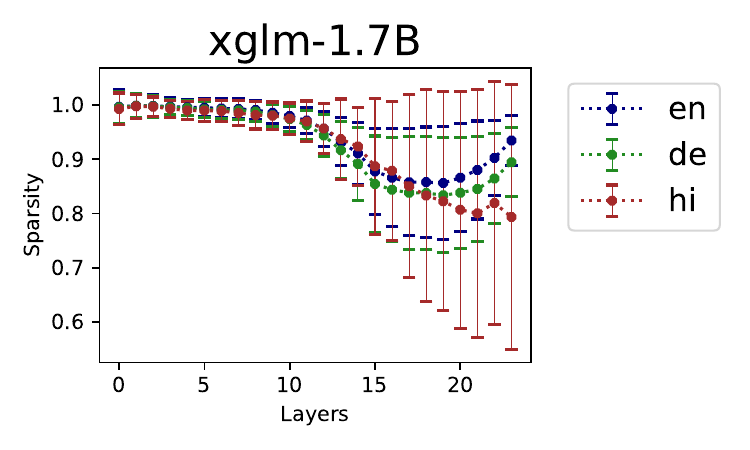}   
    \end{subfigure}
    \begin{subfigure}[b]{0.22\textwidth}
        \centering
        \includegraphics[scale=0.28]{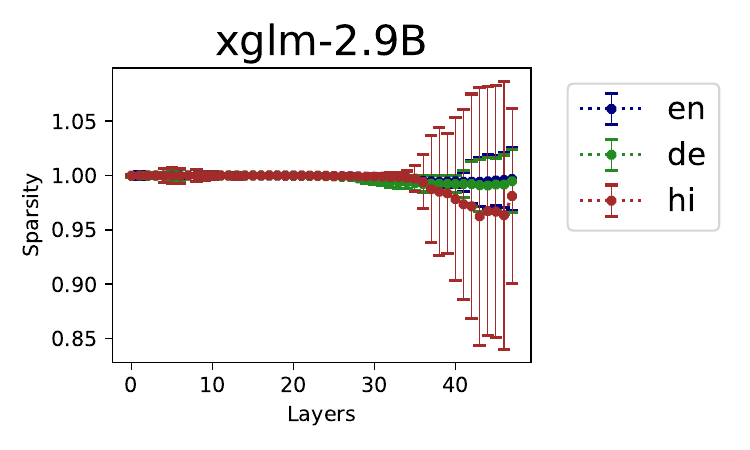}
    \end{subfigure}
    \hfill
    \begin{subfigure}[b]{0.22\textwidth}
        \centering
        \includegraphics[scale=0.28]{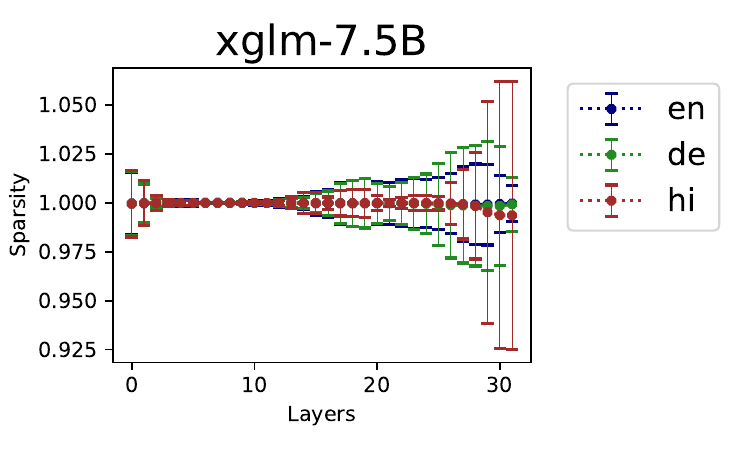}
    \end{subfigure}   
    \caption{Activation frequency for combinators along with standard deviation plotted for English, German and Hindi.}
    \label{fig:com_act_freq}
\end{figure}

For the combinators (\Cref{fig:com_act_freq}), we see a similar pattern, i.e. representations are sparser close to the input and output than the middle layers.

The authors of the XGLM model claim that the training data was curated to reflect a balanced representation of languages. Yet, it appears that lesser-represented languages (in terms of total tokens in the training data) like Czech and Hindi exhibit more combinator sparsity while well-represented languages like English, French and German exhibit greater detector sparsity in the later layers. 

One possible interpretation for this observation is that owing to their relative under-representation in the training data, languages like Czech and Hindi learn to utilise only a subset of the total neurons in the combinators for generation (greater sparsity). We therefore posit that over-sampling under-resourced languages does not improve the model's capabilities to generate tokens in those languages.

\subsection{Activation flatness across layers}
\label{activation_flatness}
Do neurons in FFNs show \textbf{similar activation patterns} across languages? How does the distribution of the values of activation change across layers? Are they more \emph{peaked} for certain languages in some layers? To answer questions like these, we make use of a novel metric called ``activation flatness''. We describe the metric and the observations from using it, in the section below.

\paragraph{Activation flatness} 
We define the activation flatness for a particular layer as :
\begin{equation}
    A = \sum_{i=1}^{n} flatness(X_i)
\end{equation}
where
\begin{equation}
    flatness(X) = \frac{-\sum_{i=1}^{m} S(x_i) \log_{2}(S(x_i))}{m}
\end{equation}
Here $m$ corresponds to the number of neurons in the layer and $S(X)$ is the neuron activation value scaled linearly to the range [0,1] within the layer, formally defined as:
\begin{equation}
    S(x_i)=\frac{x_i-min(x_1,x_2,...,x_n)}{max(x_1,x_2,...,x_n)-min(x_1,...,x_n)}
\end{equation} 

At a high level, activation flatness measures the entropy of normalized neuron activations in a layer. If all neurons return similar values, the entropy will be high, making our measure of flatness high. If only a handful of neurons fire, the entropy of these activations will be low. Thus, the activation flatness measures if the activations for a particular layer are more \emph{peaked} or \emph{uniform}. Lower flatness indicates that the activations were more peaked at a few neurons than layers with higher flatness. 

\begin{figure}[ht]  
    \centering
    \begin{subfigure}[b]{0.2\textwidth}
        \centering
        \includegraphics[scale=0.22]{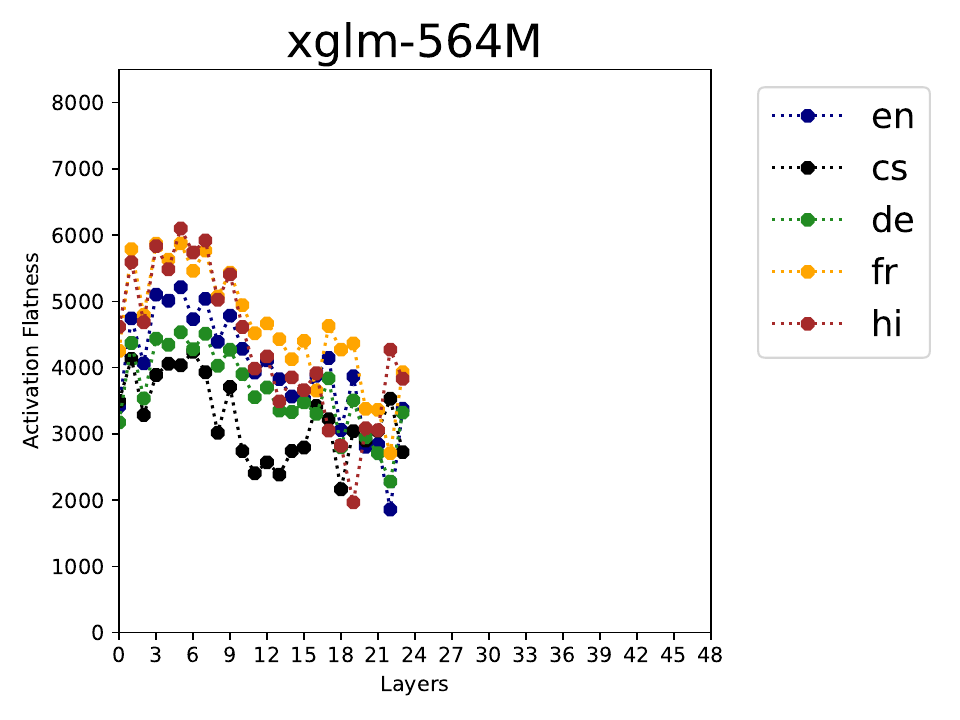}
    \end{subfigure}
    \hfill 
    \begin{subfigure}[b]{0.2\textwidth}
        \centering
        \includegraphics[scale=0.22]{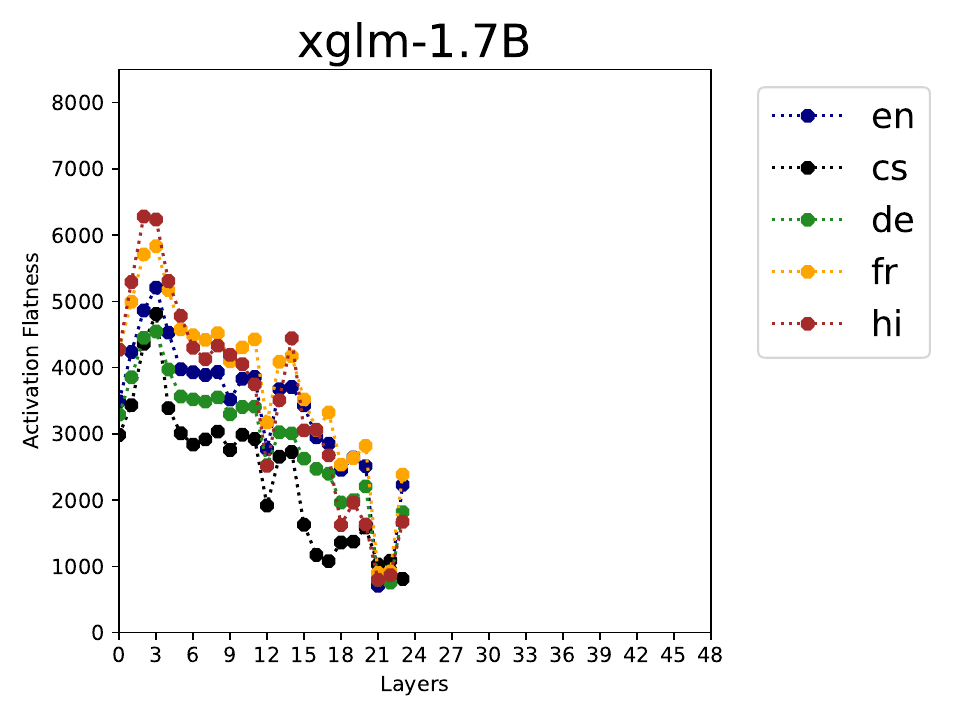}   
    \end{subfigure}
    \begin{subfigure}[b]{0.2\textwidth}
        \centering
        \includegraphics[scale=0.22]{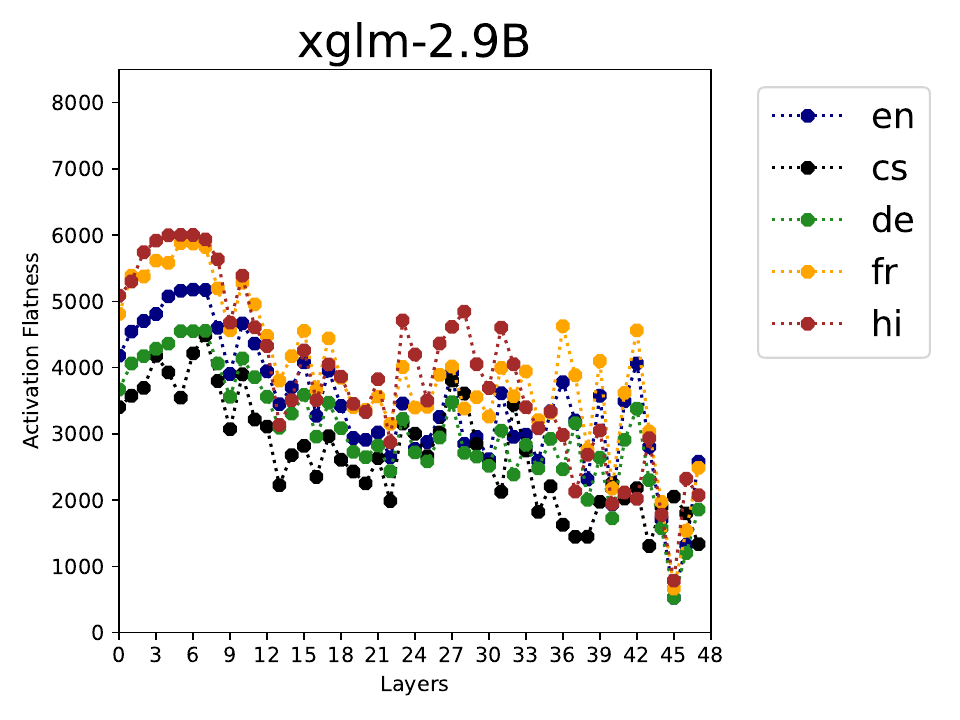}
    \end{subfigure}
    \hfill
    \begin{subfigure}[b]{0.2\textwidth}
        \centering
        \includegraphics[scale=0.22]{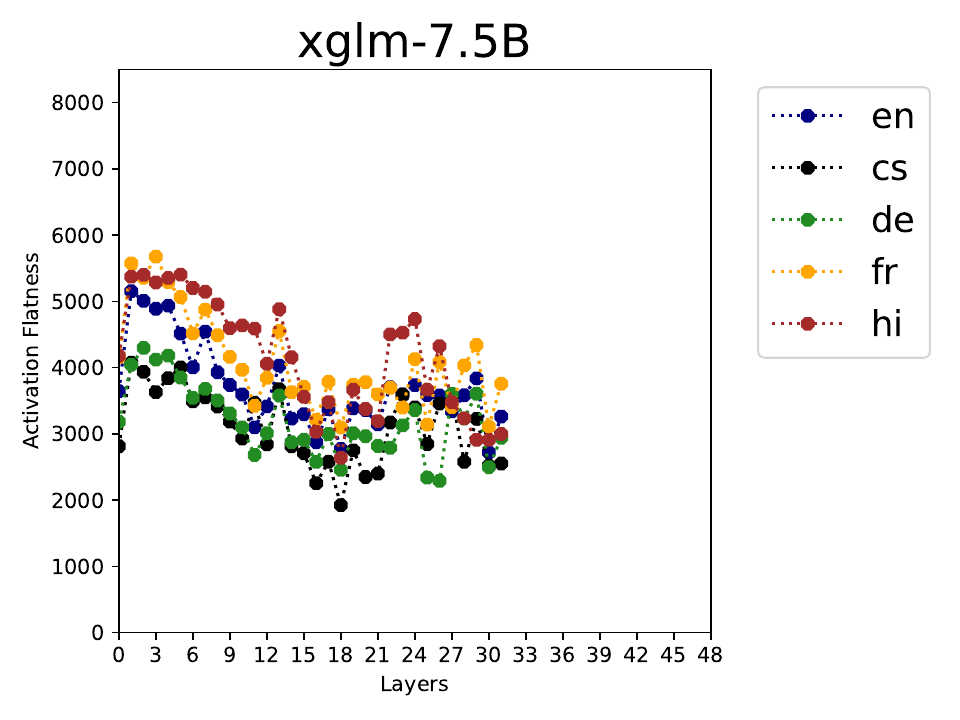}
    \end{subfigure}   
    \caption{Activation flatness for detectors}
    \label{fig:det_flatness}
\end{figure}

\begin{figure}[ht]  
    \centering
    \begin{subfigure}[b]{0.2\textwidth}
        \centering
        \includegraphics[scale=0.22]{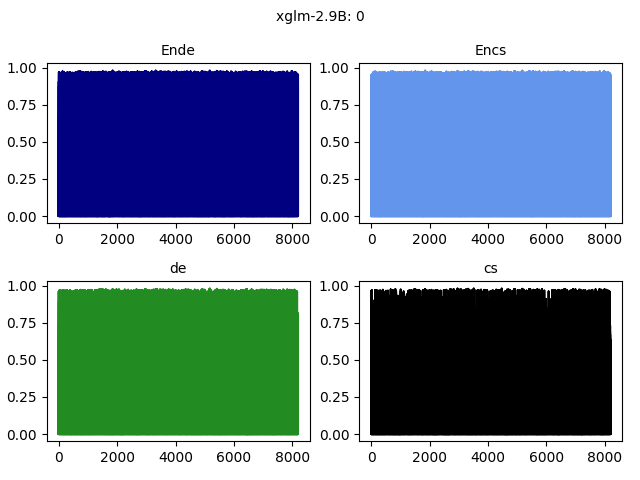}
    \end{subfigure}
    \hfill 
    \begin{subfigure}[b]{0.2\textwidth}
        \centering
        \includegraphics[scale=0.22]{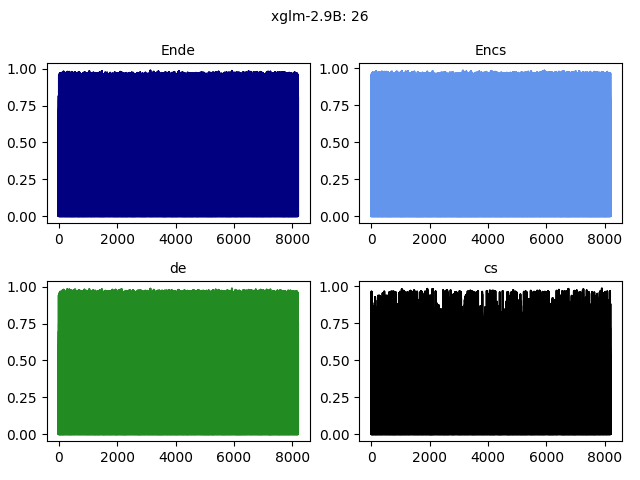}   
    \end{subfigure}
    \begin{subfigure}[b]{0.2\textwidth}
        \centering
        \includegraphics[scale=0.22]{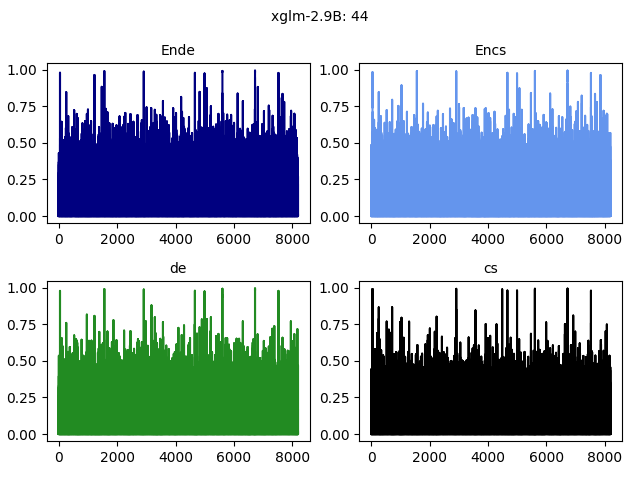}
    \end{subfigure}
    \hfill
    \begin{subfigure}[b]{0.2\textwidth}
        \centering
        \includegraphics[scale=0.22]{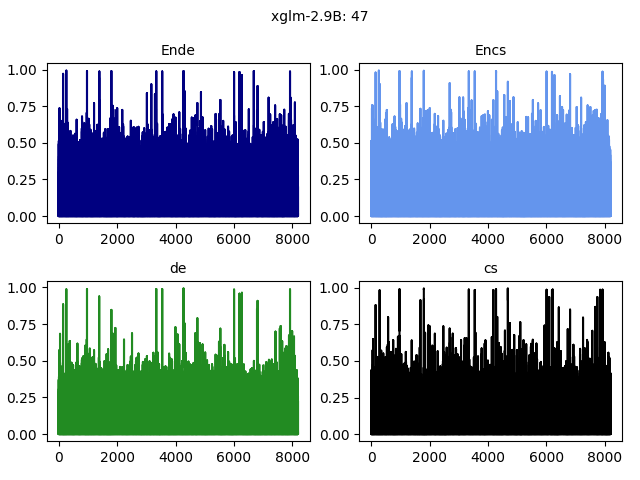}
    \end{subfigure}   
    \caption{Normalized activations for detectors: XGLM 2.9B for layers 1,27,45,48}
    \label{fig:det_norm_act}
\end{figure}

The activation flatness pattern for detectors (\Cref{fig:det_flatness}) shows a consistent pattern of decreasing flatness through the layers across all models. In other words, the activations in the detectors get more and more peaked with increasing layer depth. We see this clearly in \Cref{fig:det_norm_act} where we look at the activations across all prefixes for detectors of some layers. The activation patterns for layers 1 and 27 (uniform), are for instance very different from those of layers 45 and 48 (peaked at certain neurons). 

\begin{figure}[ht]  
    \centering
    \begin{subfigure}[b]{0.2\textwidth}
        \centering
        \includegraphics[scale=0.22]{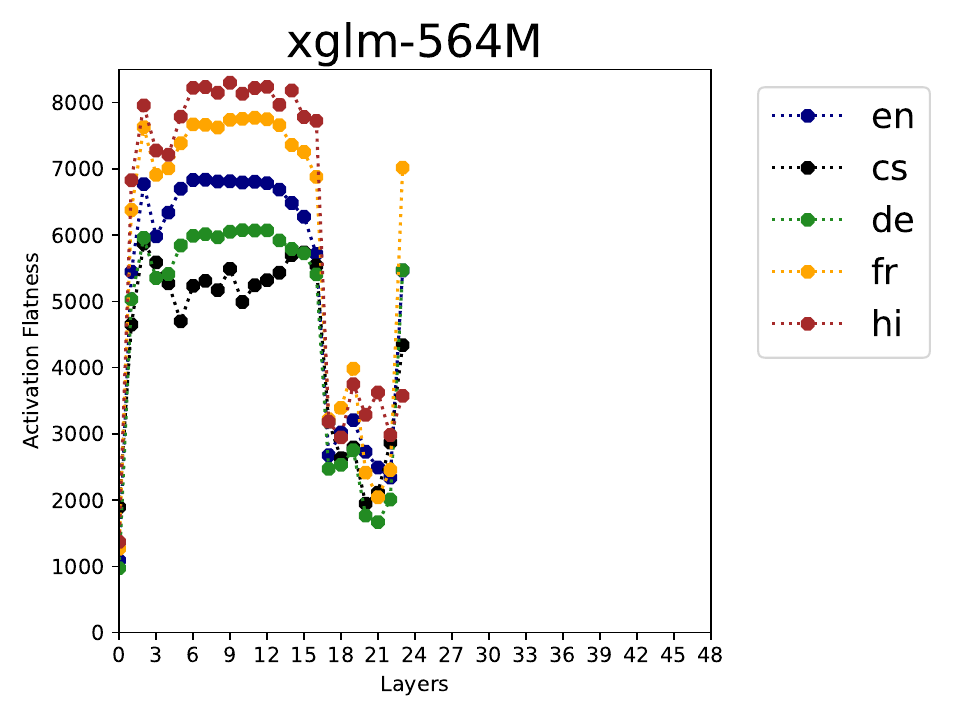}
    \end{subfigure}
    \hfill 
    \begin{subfigure}[b]{0.2\textwidth}
        \centering
        \includegraphics[scale=0.22]{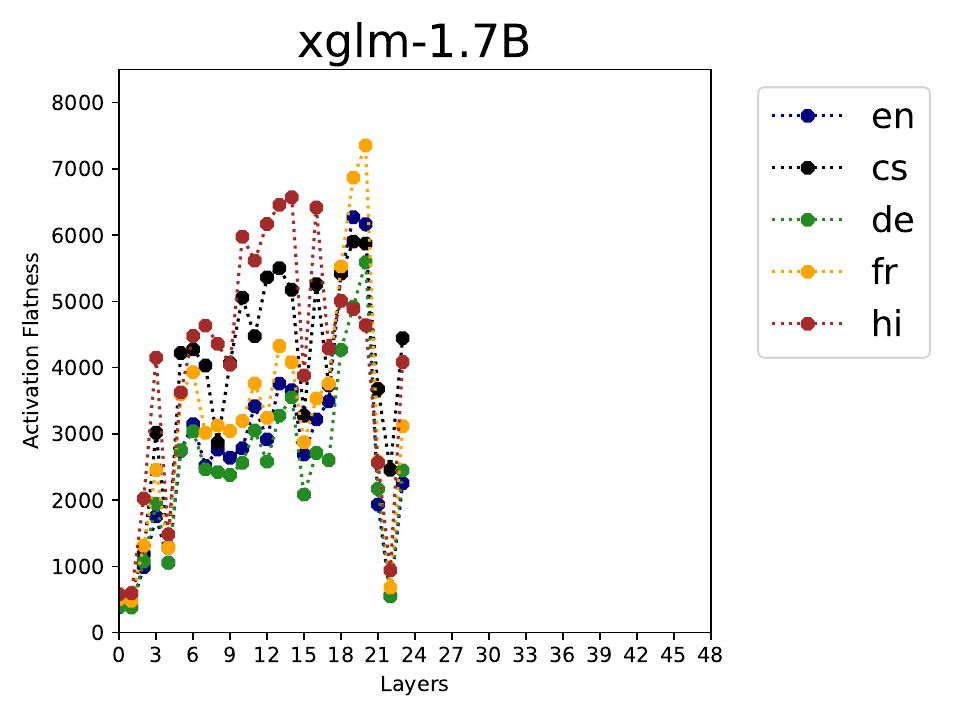}   
    \end{subfigure}
    \begin{subfigure}[b]{0.2\textwidth}
        \centering
        \includegraphics[scale=0.22]{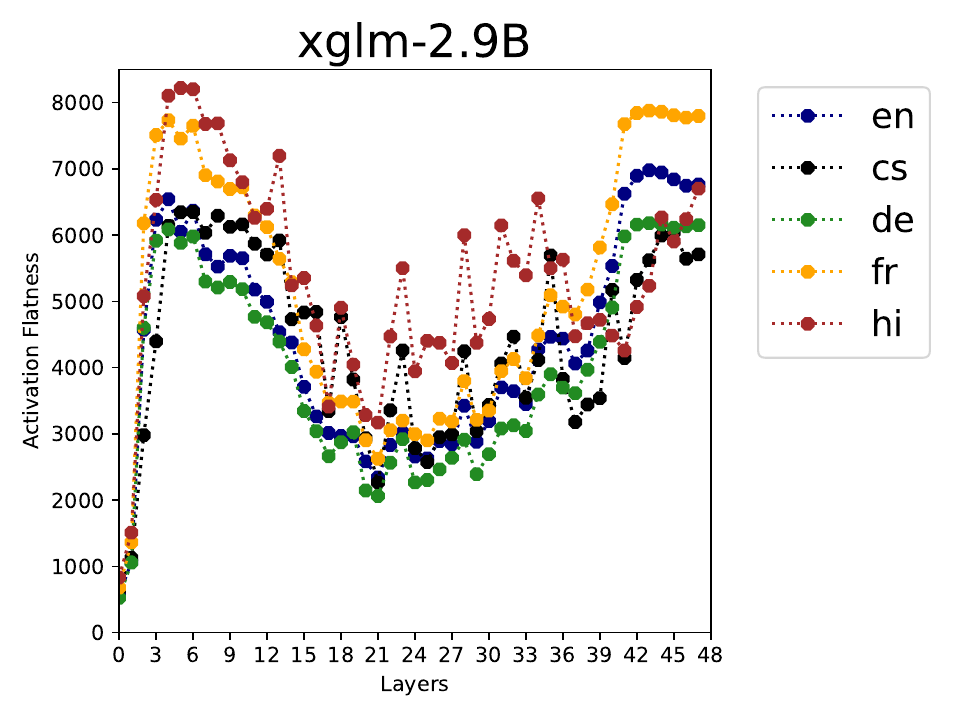}
    \end{subfigure}
    \hfill
    \begin{subfigure}[b]{0.2\textwidth}
        \centering
        \includegraphics[scale=0.22]{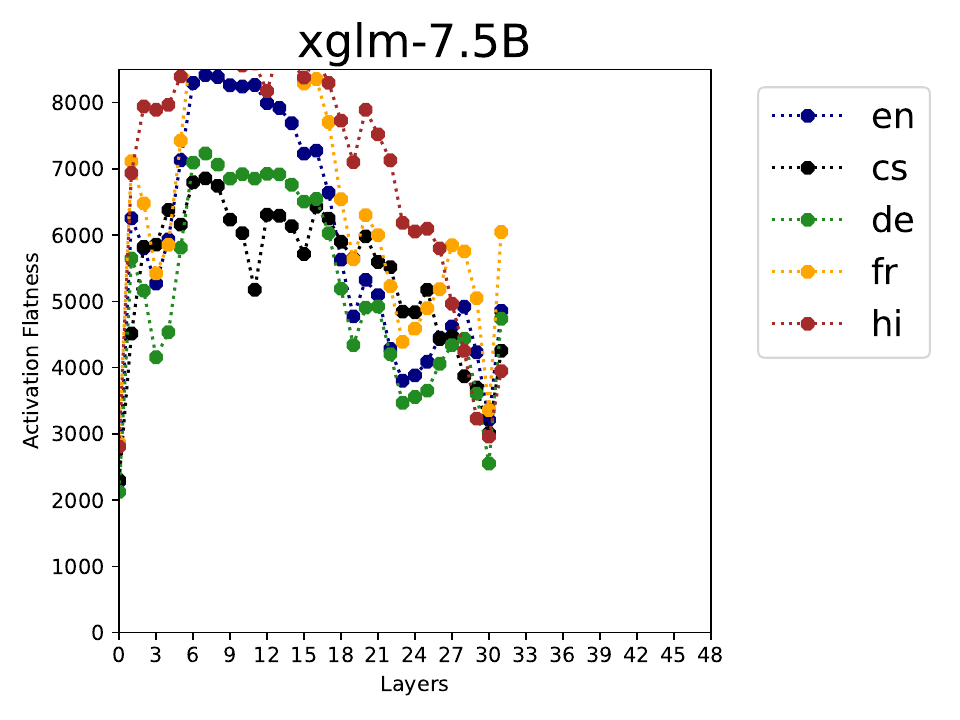}
    \end{subfigure}   
    \caption{Activation flatness for combinators}
    \label{fig:com_flatness}
\end{figure}

\begin{figure}[ht]  
    \centering
    \begin{subfigure}[b]{0.2\textwidth}
        \centering
        \includegraphics[scale=0.22]{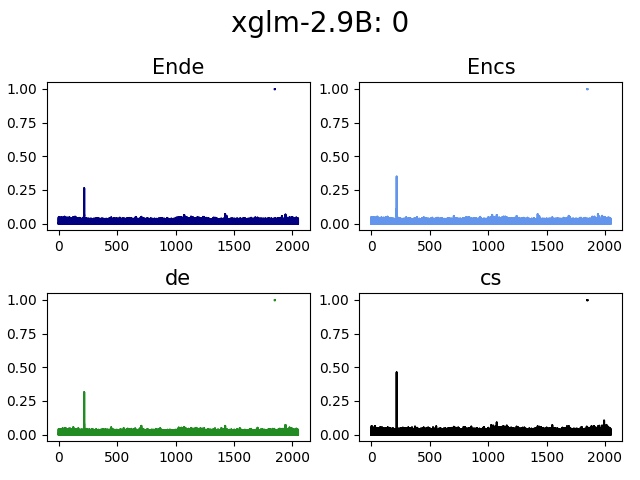}
    \end{subfigure}
    \hfill 
    \begin{subfigure}[b]{0.2\textwidth}
        \centering
        \includegraphics[scale=0.22]{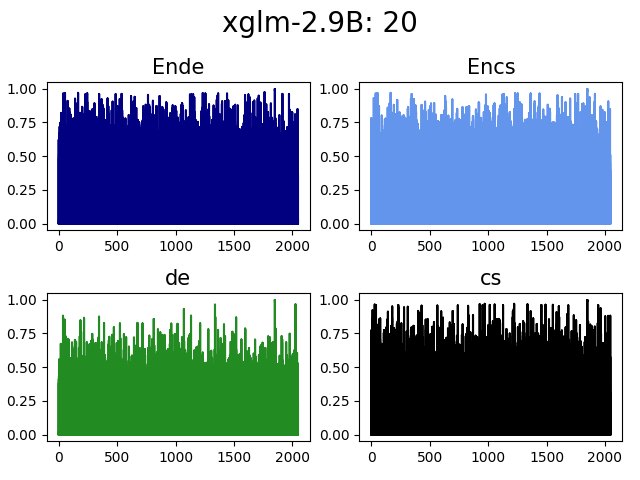}   
    \end{subfigure}
    \begin{subfigure}[b]{0.2\textwidth}
        \centering
        \includegraphics[scale=0.22]{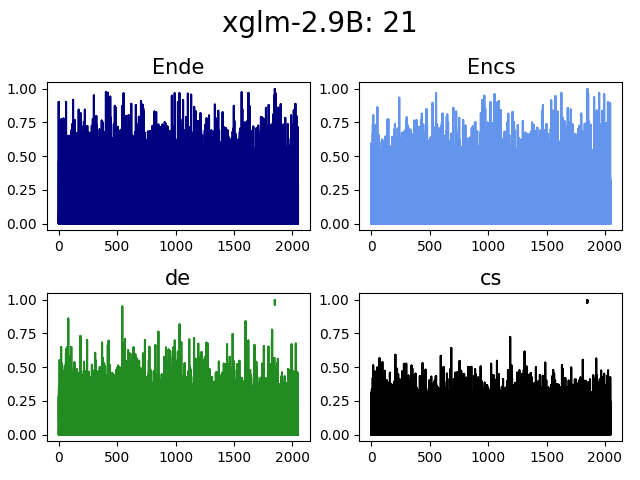}
    \end{subfigure}
    \hfill
    \begin{subfigure}[b]{0.2\textwidth}
        \centering
        \includegraphics[scale=0.22]{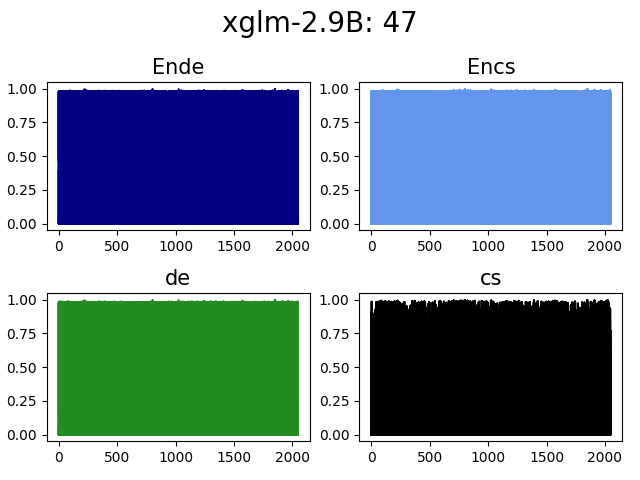}
    \end{subfigure}   
    \caption{Normalized activations for combinators: XGLM 2.9B for layers 1, 21, 22, and 47}
    \label{fig:com_norm_act}
\end{figure}

The activation flatness pattern for combinators (\Cref{fig:com_flatness}) shows that all models follow a common pattern. Regardless of the number of parameters and the total number of layers, there is a distinct decrease in the activation flatness near layer 20 for all models. A visual inspection of \Cref{fig:com_norm_act} shows that the activations for the prefixes indeed show a pattern of certain `peakedness' in layers of low activation flatness (i.e. layers 1, 21, 22 vs. 47). Also the drop in activation flatness occurs across all languages for every model. From the detector-combinator perspective, the detector does pattern matching from the input representation. The combinator then uses the output of the detector (after being passed through an activation function) to make an intermediate prediction for that layer. Any drop in the activation flatness for combinators thus indicates more `focused' firing along specific dimensions.

\begin{figure}[ht]  
    \centering
    \begin{subfigure}[b]{0.5\textwidth}
        \centering
        \includegraphics[scale=0.28]{images/combinator_pics/2.9/0.png}
    \end{subfigure}
    \hfill 
    \begin{subfigure}[b]{0.5\textwidth}
        \centering
        \includegraphics[scale=0.28]{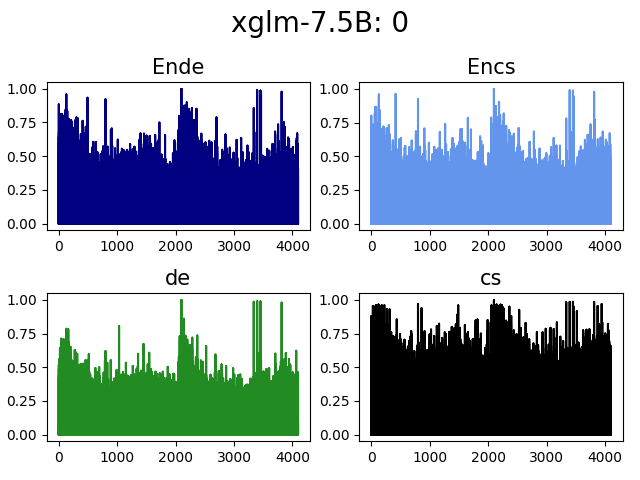}
    \end{subfigure}
    \caption{Combinator activation for all prefixes for layer 1: 2.9B (top, activation flatness value of $\sim$1000) vs. 7.5B (bottom, activation flatness of $\sim$2000) models.}
    \label{fig:com_1_act}
\end{figure}

To better illustrate how activation flatness works, we also compare the activation snapshots for layer 1 of the 2.9B and 7.5B parameter model in \Cref{fig:com_1_act}. We observe that for the first layer of the 2.9B parameter model, where the combinator activation flatness value was around 1000 for all languages, there is a single distinct peak in the activation distribution compared to the first layer of the 7.5B parameter model (with activation flatness of around 2000 across languages). Upon closer inspection we find that for the first layer of the 2.9B model, the maximum activation is always recorded for neuron 1849 for every prefix across all languages. Also across all languages, the maximum standard deviation across prefixes was recorded for neuron 218 for all languages. The single peak that we observe for the 2.9B model in \Cref{fig:com_1_act} corresponds to the neuron 218. Thus, the aggregate value of our activation flatness reflects the intuitive differences in representations.

For both detectors and combinators, we observe that the low values of activation flatness occurs similarly across all languages. We therefore investigate the extent to which the prefixes across languages exhibit similarity in their layer-wise representations. Representational similarity would point to the presence of \textbf{multilingual neurons} in those layers. 

\begin{algorithm}[htbp]
  \caption{Representation Distance Score}
  \label{alg:min_distance_score}
  \SetAlgoLined
  \KwIn{Layer snapshot $L_1$ with shape $(m, d)$ and snapshot $L_2$ with shape $(n, d)$}
  \KwOut{Aggregate minimum distance across all rows of $L_1$, i.e. all prefixes of the source sentence and their most similar counterpart prefixes in the target language}
  $total\_dist \gets 0 $\;
  \For{$i \gets 1$ \KwTo $m$}{
    $min\_dist \gets +\infty$;\\
    \For{$j \gets 1$ \KwTo $n$}{
        $score \gets dist({L_1[i],L_2[j]}$;\\
        \If{$score < min\_dist$}{
                $min\_dist \gets score$ 
            }
    }
    $total\_dist \gets total\_dist + min\_dist$
  }
  \Return{$total\_dist$}
\end{algorithm}

We posit that if a layer has more multilingual neurons, then the representation of two sentence prefixes with the same meaning should be similar in that layer. In other words, the representation of ``I am a computer'' should be very similar to ``Ja jsem počítač'' (in Czech). Thus for every prefix in one language, we find the representational difference with all prefixes in the second language. We then select the minimum distance among them. The idea is to identify `similar' prefixes across both languages. For simplicity, we assume monotonic translation, which more or less holds for our studied language pairs.  We then aggregate the distances across all prefixes. The layer with the minimal aggregate distance should exhibit the greatest representational similarity for the two languages, as formally captured in \Cref{alg:min_distance_score}.  

\begin{figure}[ht]  
    \centering
    \begin{subfigure}[b]{0.2\textwidth}
        \centering
        \includegraphics[scale=0.22]{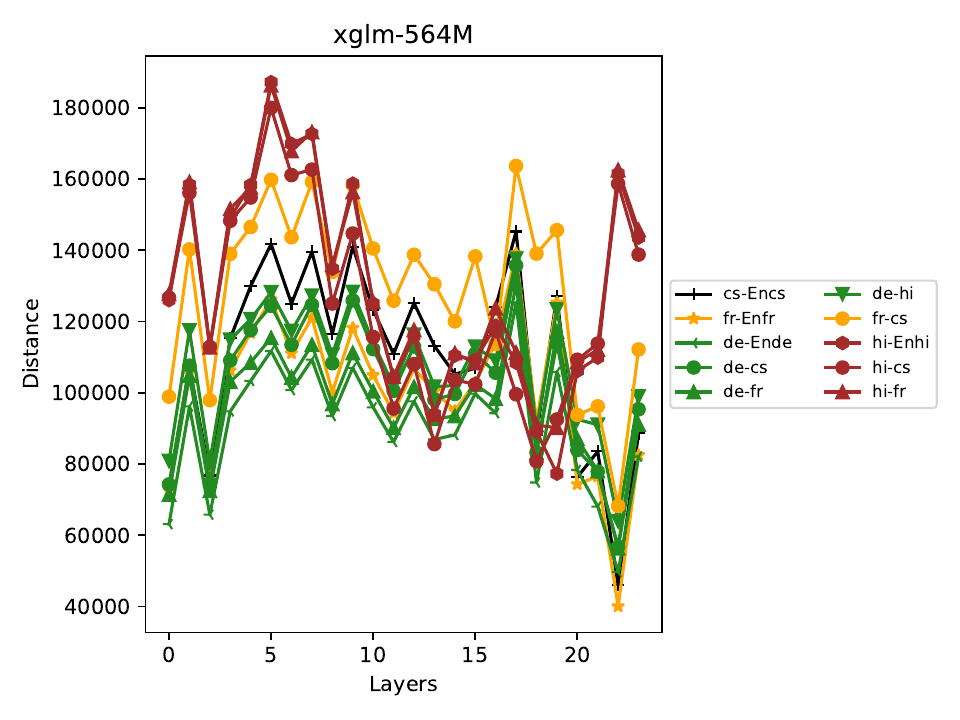}
    \end{subfigure}
    \hfill 
    \begin{subfigure}[b]{0.2\textwidth}
        \centering
        \includegraphics[scale=0.22]{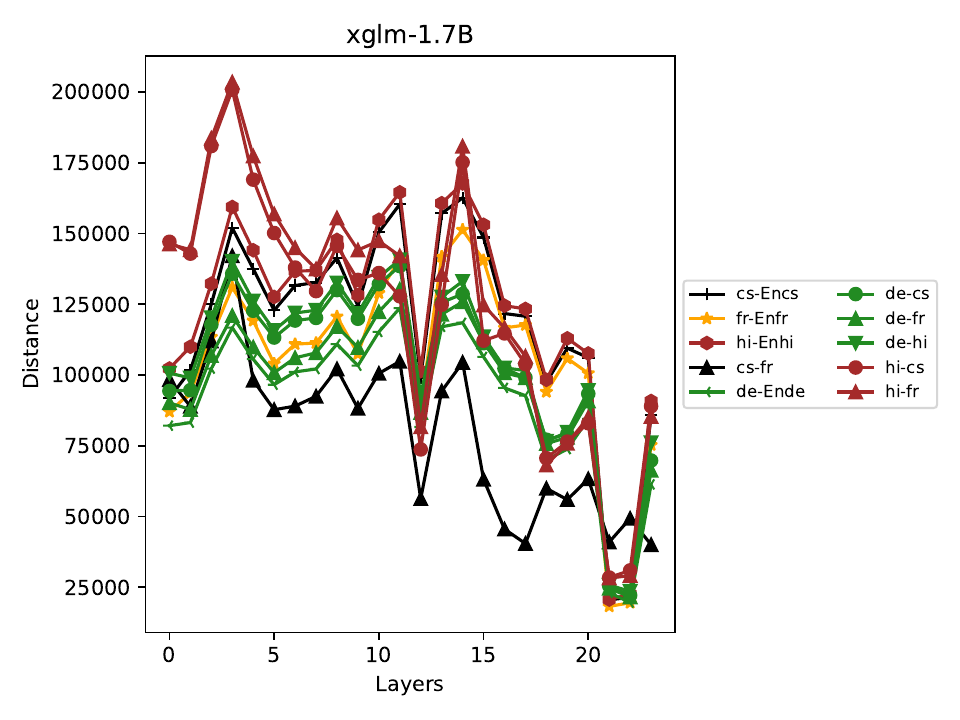}
    \end{subfigure}
    \begin{subfigure}[b]{0.2\textwidth}
        \centering
        \includegraphics[scale=0.22]{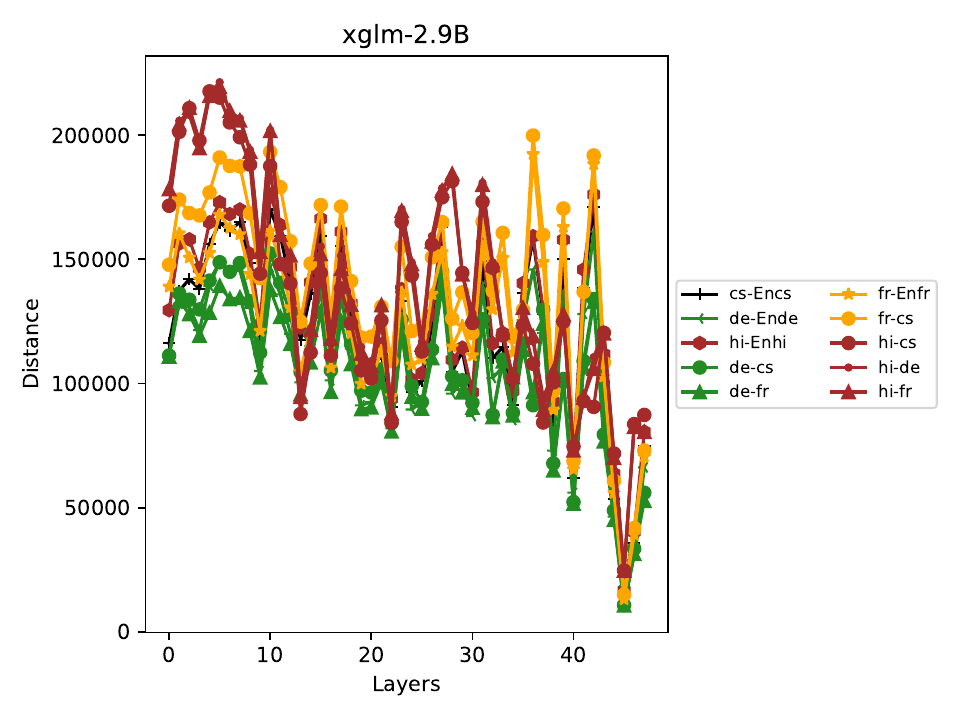}
    \end{subfigure}
    \hfill
    \begin{subfigure}[b]{0.2\textwidth}
        \centering
        \includegraphics[scale=0.22]{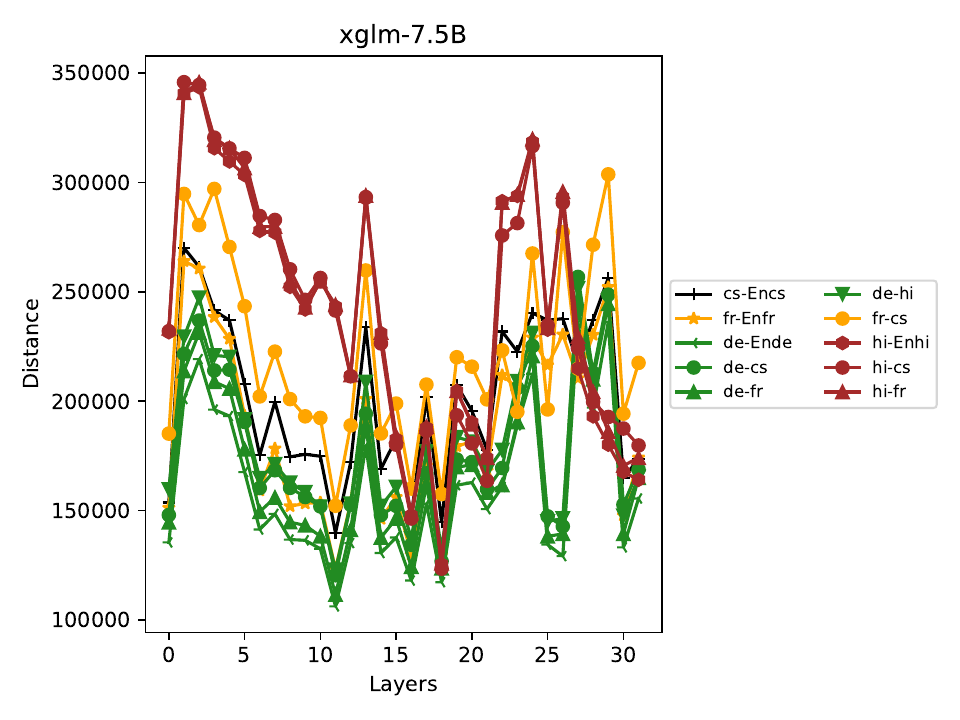}
    \end{subfigure}   
    \caption{Distance between representations: Detectors (568M, 1.7B. 2.9B, 7.5B)}
    \label{fig:det_dist}
\end{figure}

\begin{figure}[htb]  
    \centering
    \begin{subfigure}[b]{0.2\textwidth}
        \centering
        \includegraphics[scale=0.22]{images//tensor_plots/xglm-564M_detectors.pdf}
    \end{subfigure}
    \hfill 
    \begin{subfigure}[b]{0.2\textwidth}
        \centering
        \includegraphics[scale=0.22]{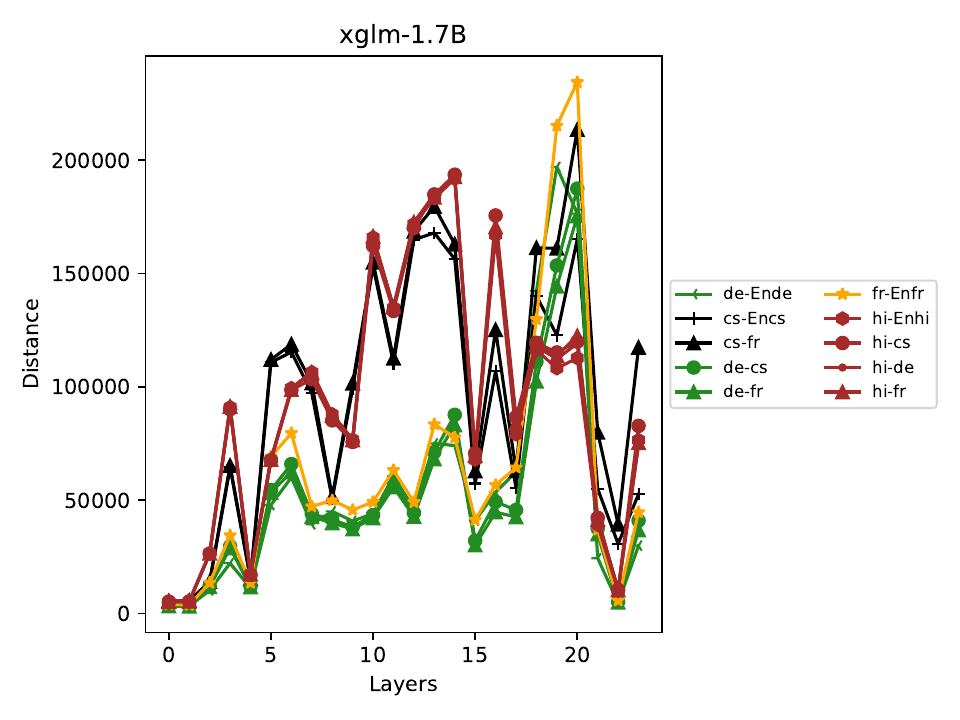}
    \end{subfigure}
    \begin{subfigure}[b]{0.2\textwidth}
        \centering
        \includegraphics[scale=0.22]{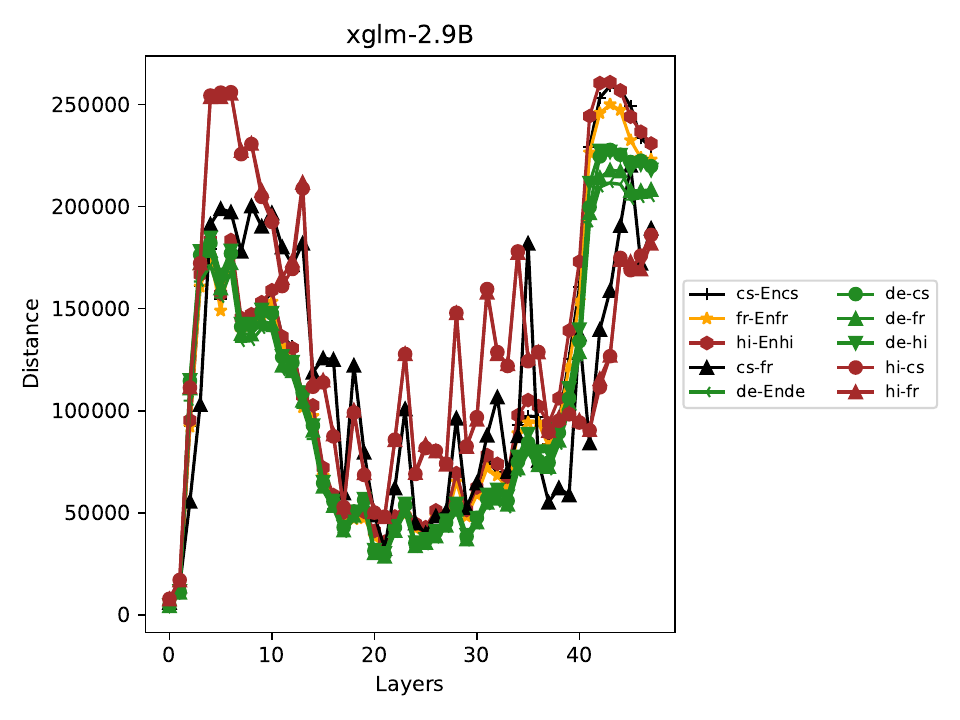}
    \end{subfigure}
    \hfill
    \begin{subfigure}[b]{0.2\textwidth}
        \centering
        \includegraphics[scale=0.22]{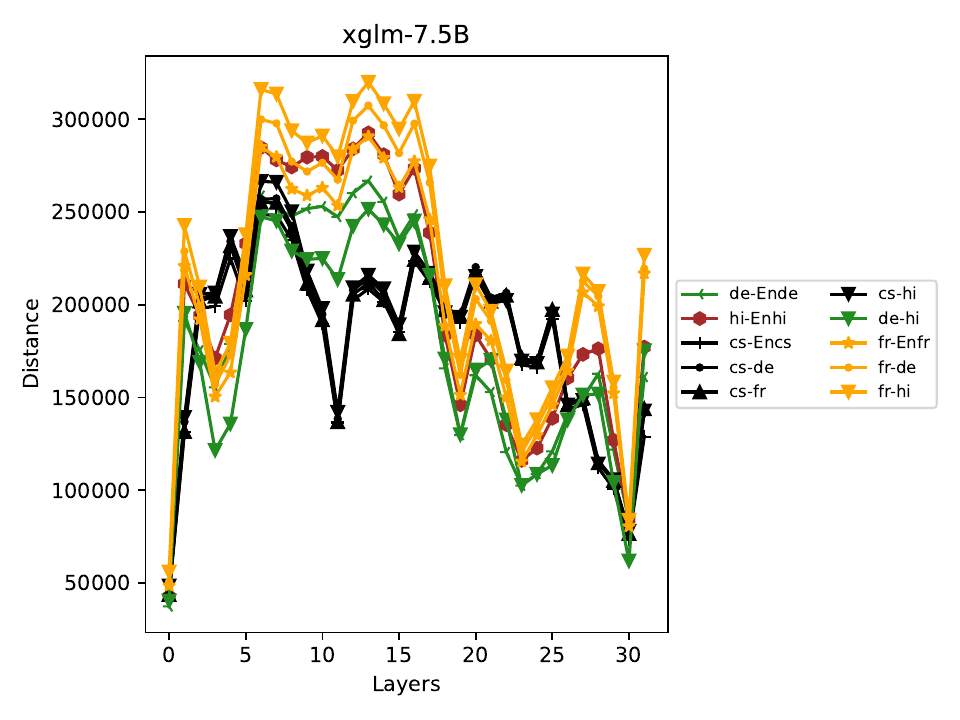}
    \end{subfigure}   
    \caption{Distance between representations: Combinators (568M, 1.7B. 2.9B, 7.5B)}
    \label{fig:com_dist}
\end{figure}

From \Cref{fig:det_dist} we see that the representational distance for detectors decreases with layer depth for all models except the 7.5B parameter model. \Cref{fig:com_dist} shows that for the 2.9B parameter model, the representational distance for combinators decreases in the middle, before increasing again. Interestingly, the pattern of detector representational distance for 7.5B seems similar to the pattern of combinator representational distance for the 2.9B model. We also observe that around layer 30 in the 7.5B parameter model, the representational distance drops significantly for the combinators before increasing abruptly through layers 31 and 32. We label the layers with the minimal distance between prefixes of different language pairs as relatively more \emph{multilingual}.

\begin{figure}[ht]  
    \centering
    \begin{subfigure}[b]{0.2\textwidth}
        \centering
        \includegraphics[scale=0.23]{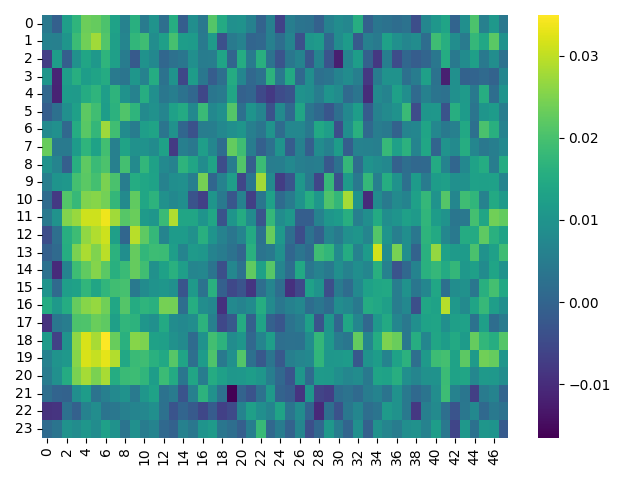}
    \end{subfigure}
    \hfill 
    \begin{subfigure}[b]{0.2\textwidth}
        \centering
        \includegraphics[scale=0.23]{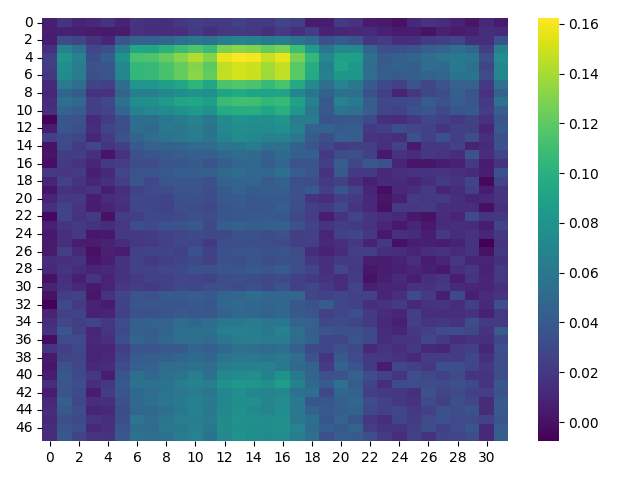}   
    \end{subfigure}
    \hfill 
    \begin{subfigure}[b]{0.2\textwidth}
        \centering
        \includegraphics[scale=0.23]{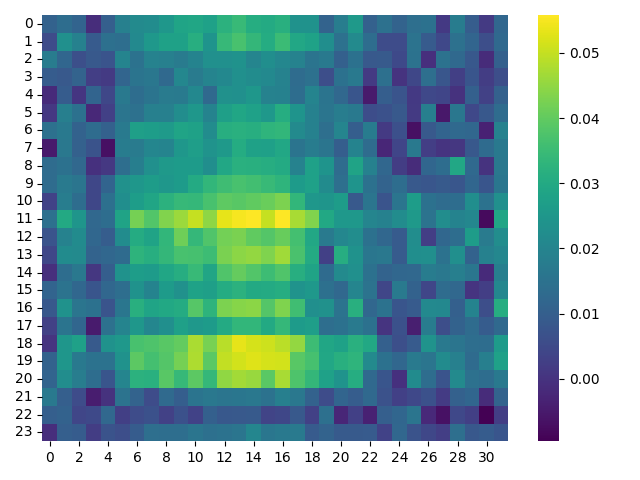}   
    \end{subfigure}
    \caption{Representational Similarity between models (combinators): 1.7B vs 2.9B, 2.9B vs 7.5B, 1.7B vs 7.5B}
    \label{fig:rep_sim}
\end{figure}

Finally, we also compare the ``representational similarity'' \cite{kriegeskorte2008representational} for German between the 1.7B, 2.9B and 7.5B parameter models in \Cref{fig:rep_sim}. We calculate the representational similarity on the basis of the normalized values of activation corresponding to each prefix for the detectors/combinators of all layers. We see that the similarity is the maximum for 2.9B and 7.5B models. Particularly, some of the middle layers of the 7.5B parameter model shows high correlation with early and middle layers of the 2.9B parameter model.    

The representational distance for all language pairs at the last layers of the combinator is higher for the models with larger parameters. In other words, it appears that the final layer of the larger models can make distinction between different languages. We posit that this might be an explanation why the 7.5B model outperforms models with smaller parameters of the same family as reported by \citet{lin2022few}.

Thus, based on the observations of activation flatness and representational distances/similarity, we have a comprehensive view of the nature of multilinguality in detectors and combinators. For all models except 7.9B, the detectors keep getting more multilingual with layer depth. For the 7.5B model, the middle layers are more multilingual than the layers near the input/output. For the combinators, final layers are the most multilingual except the 2.9B parameter model where the middle layers are the most multilingual.

\subsection{The strange behaviour of the 2.9B model}
\label{strange_2.9B}

We find that the combinators of the 2.9B billion parameter behave very differently through the layers than the combinators of the other models. In terms of the architectural micro-details, model dimension size and the hidden dimension size of the 2.9B model is the same as that of the 1.7B model i.e. 2048 and 8192 respectively. The 2.9B model differs from the 1.7B model only in terms of layer depth i.e. 24 vs 48. There is also representational similarity across layers of the two models. We see that for both models, there is a minimum in activation flatness (as well as minimum representational distance) near layer 20. For the 2.9B billion parameter model, we also observe that the representational distance gradually increases in the later layers. For the final layer, we also find that for both the 7.5B and 2.9B models, the representational distance among language pairs is almost comparable ($\sim$20000). One major point of difference between the 2.9B and 7.5B parameter model (except the architectural micro-details) is the number of layers in the model. 

We suspect that this might make the 2.9B model bad at generation (in comparison to the 7.5B model). We thus posit that the increase in number of layers harms the generation quality of the 2.9B parameter model \textbf{(over-layerization)}. We suspect that longer training of the 2.9B model could have fixed this but we have no further details about the training of XGLM.

The authors of the XGLM report that the 7.5B parameter model outperforms other smaller models of the same family in mutilingual tasks including few-shot translation. However, recent work by \citet{zhang2023machine} involving QLoRA fine tuning of LLMs for translation shows that after finetuning, the XGLM 2.9B parameter model outperforms larger XGLM models in translation tasks. Is this boost in performance somehow related to the activation flatness patterns observed by us? We plan to investigate this aspect in future work, hypothesizing that flatter activation patters could be easier to adapt in QLoRA.

\subsection{Activation similarity through languages}
We have already seen some evidence that certain layers are more multilingual than others based on our metric of representational distance/similarity. In this section, we validate the observations and confirm if the same neurons process different languages in the layers. 

For the model snapshots of detectors and combinators from each layer, we take the sum of activations across all prefixes and rank the neurons based on that. The motivation here is that an aggregate picture might reveal which neurons `light up' the most across prefixes. 

We then calculate the Spearman rank correlation for a layer between language pairs. Our hypothesis posits that high rank correlations between two languages within a layer suggest similarities in their processing patterns. We present the results below.   

\begin{figure}[ht]  
    \centering
    \begin{subfigure}[b]{0.2\textwidth}
        \centering
        \includegraphics[scale=0.24]{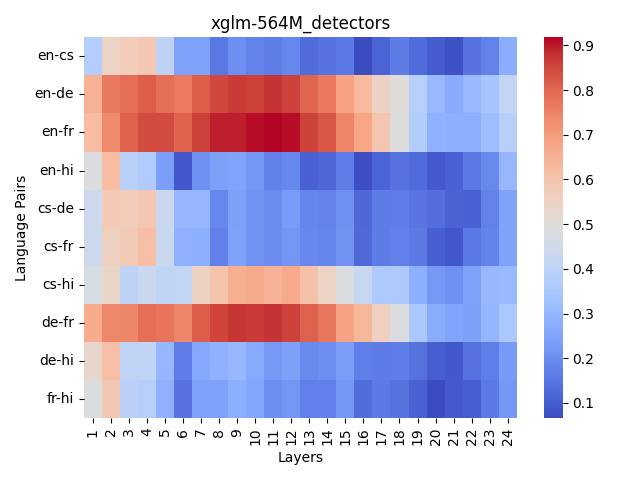}
    \end{subfigure}
    \hfill 
    \begin{subfigure}[b]{0.2\textwidth}
        \centering
        \includegraphics[scale=0.24]{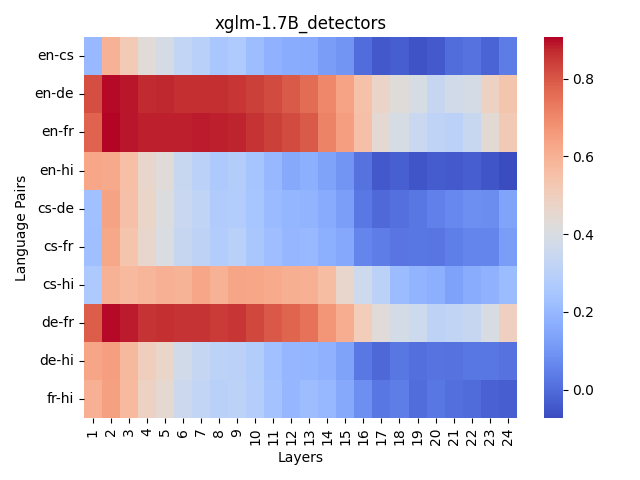}   
    \end{subfigure}
    \begin{subfigure}[b]{0.2\textwidth}
        \centering
        \includegraphics[scale=0.24]{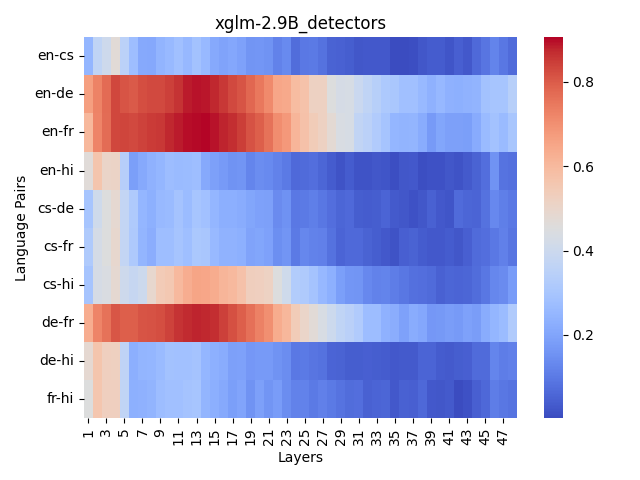}
    \end{subfigure}
    \hfill
    \begin{subfigure}[b]{0.2\textwidth}
        \centering
        \includegraphics[scale=0.24]{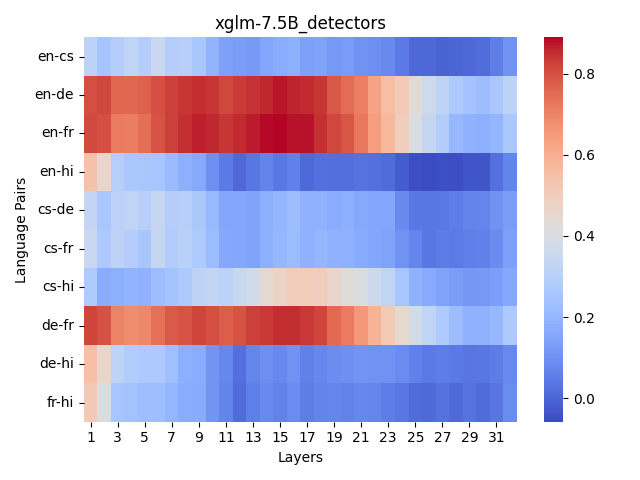}
    \end{subfigure}   
    \caption{Rank correlation of neurons in detectors}
    \label{fig:det_sim}
\end{figure}

\begin{figure}[ht]  
    \centering
    \begin{subfigure}[b]{0.2\textwidth}
        \centering
        \includegraphics[scale=0.24]{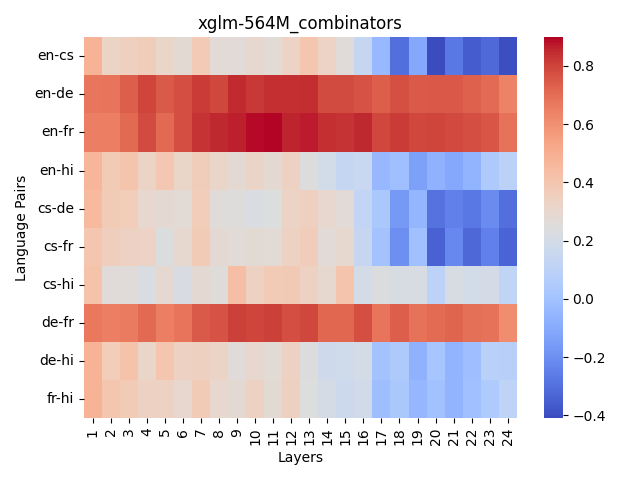}
    \end{subfigure}
    \hfill 
    \begin{subfigure}[b]{0.2\textwidth}
        \centering
        \includegraphics[scale=0.24]{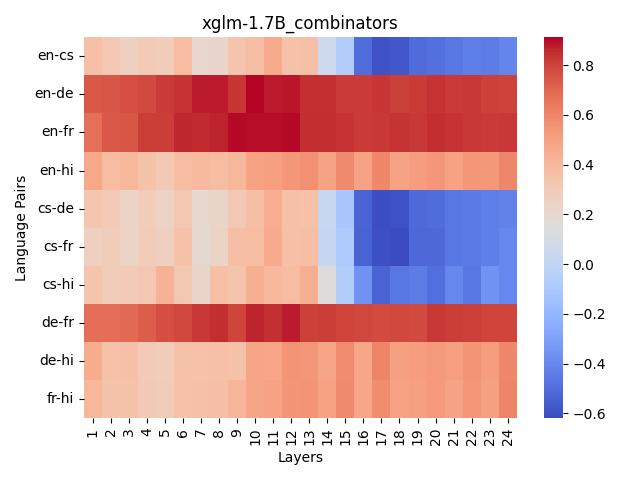}   
    \end{subfigure}
    \begin{subfigure}[b]{0.2\textwidth}
        \centering
        \includegraphics[scale=0.24]{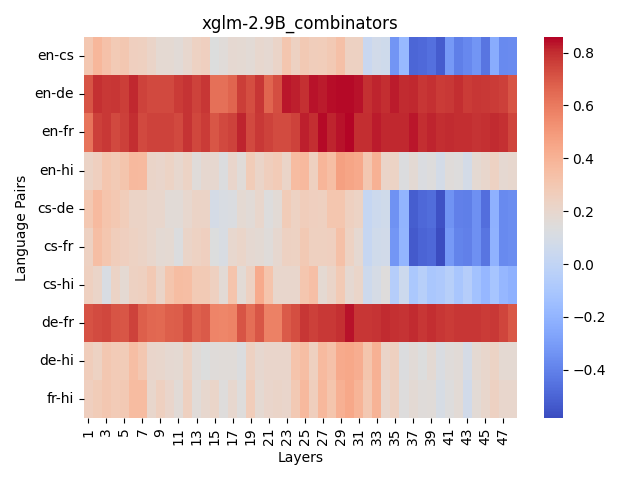}
    \end{subfigure}
    \hfill
    \begin{subfigure}[b]{0.2\textwidth}
        \centering
        \includegraphics[scale=0.24]{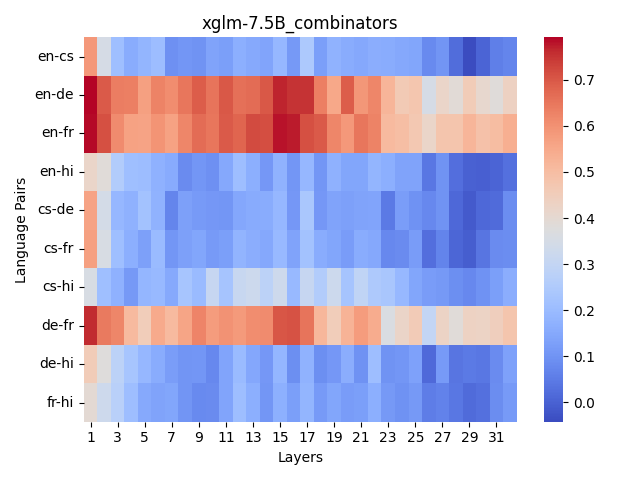}
    \end{subfigure}   
    \caption{Rank correlation of neurons in combinators}
    \label{fig:com_sim}
\end{figure}

We observe that layers of detectors (\Cref{fig:det_sim}) exhibit decreasing rank correlation across languages with increasing model size. However, the rank correlation is consistently high for German-English-French language pairs across models. From \Cref{fig:det_sim} and \Cref{fig:com_sim}, we also find that the combinators show greater rank correlation than detectors across languages and across layers, aligning again with the observations made previously with representational similarity. 

For both detectors and combiantors, we observe high degrees of correlation in the early and middle layers for all models. For the combinator we observe an \textbf{`emerging'} pattern where the final layers initially exhibit greater values of rank correlation (568M<1.7B) before decreasing again (1.7B>2.9B>7.5B). We also observe that the middle combinator layers of the 7.5B model exhibited the greatest rank correlation for the English-German-French pairs.

Thus, the early detectors are more multilingual for all models. We also observe that the early and middle combinator layers become multilingual with increasing model size.

\begin{figure}[ht]  
    \centering
    \begin{subfigure}[b]{0.2\textwidth}
        \centering
        \includegraphics[scale=0.24]{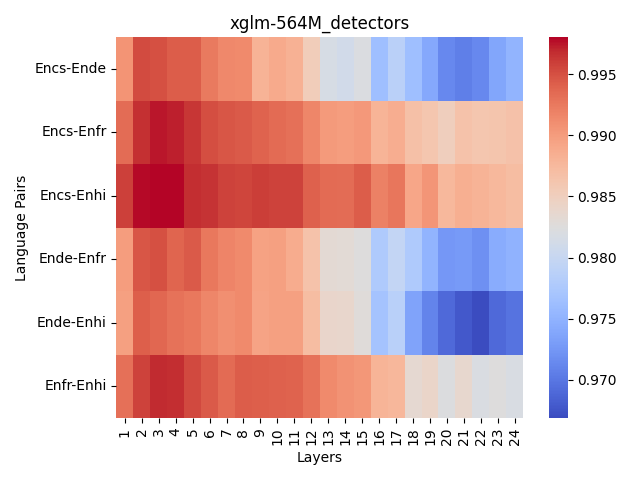}
    \end{subfigure}
    \hfill 
    \begin{subfigure}[b]{0.2\textwidth}
        \centering
        \includegraphics[scale=0.24]{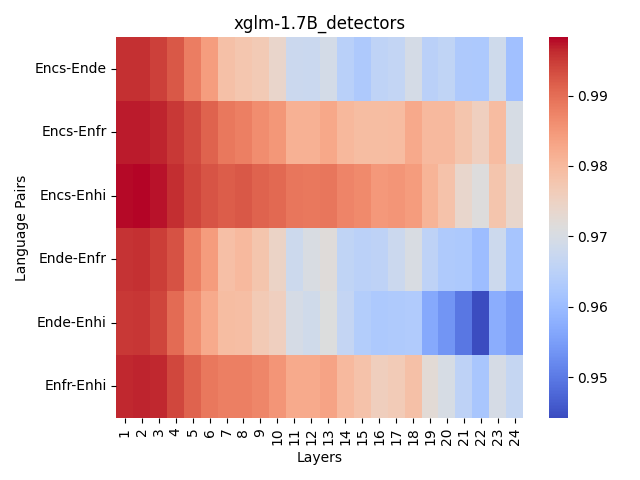}   
    \end{subfigure}
    \begin{subfigure}[b]{0.2\textwidth}
        \centering
        \includegraphics[scale=0.24]{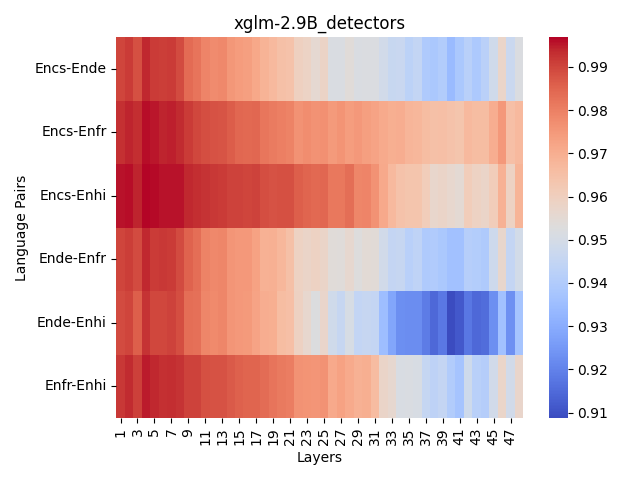}
    \end{subfigure}
    \hfill
    \begin{subfigure}[b]{0.2\textwidth}
        \centering
        \includegraphics[scale=0.24]{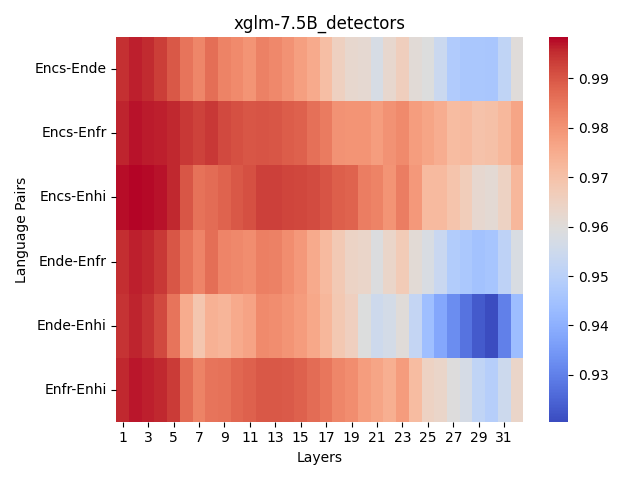}
    \end{subfigure}   
    \caption{Similarity of ranks (English) : detectors}
    \label{fig:det_sim_eng}
\end{figure}

\begin{figure}[ht]  
    \centering
    \begin{subfigure}[b]{0.2\textwidth}
        \centering
        \includegraphics[scale=0.24]{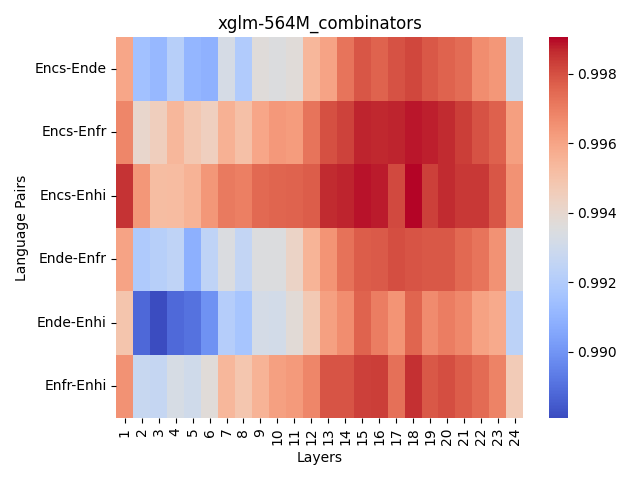}
    \end{subfigure}
    \hfill 
    \begin{subfigure}[b]{0.2\textwidth}
        \centering
        \includegraphics[scale=0.24]{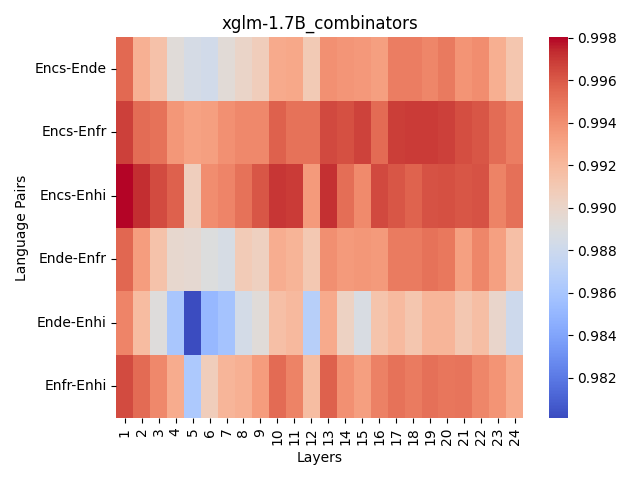}   
    \end{subfigure}
    \begin{subfigure}[b]{0.2\textwidth}
        \centering
        \includegraphics[scale=0.24]{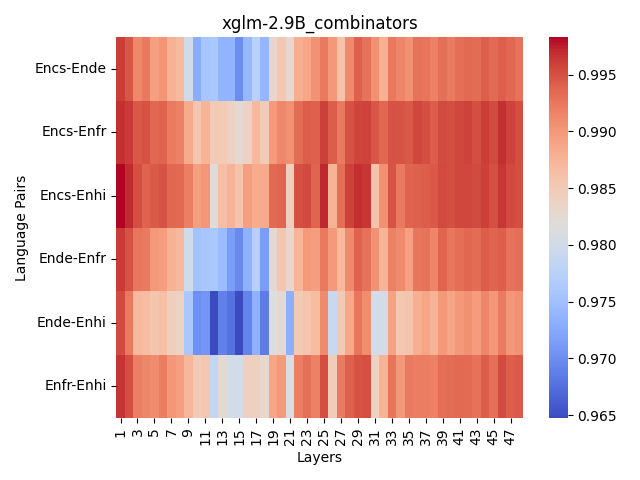}
    \end{subfigure}
    \hfill
    \begin{subfigure}[b]{0.2\textwidth}
        \centering
        \includegraphics[scale=0.24]{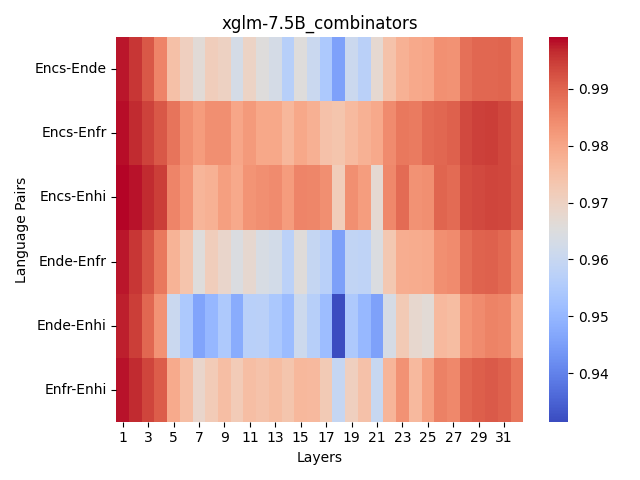}
    \end{subfigure}   
    \caption{Similarity of ranks (English) : combinators}
    \label{fig:com_sim_eng}
\end{figure}

\paragraph{Intra-language comparison}

\begin{figure}[ht]  
    \centering
    \includegraphics[scale=0.3]{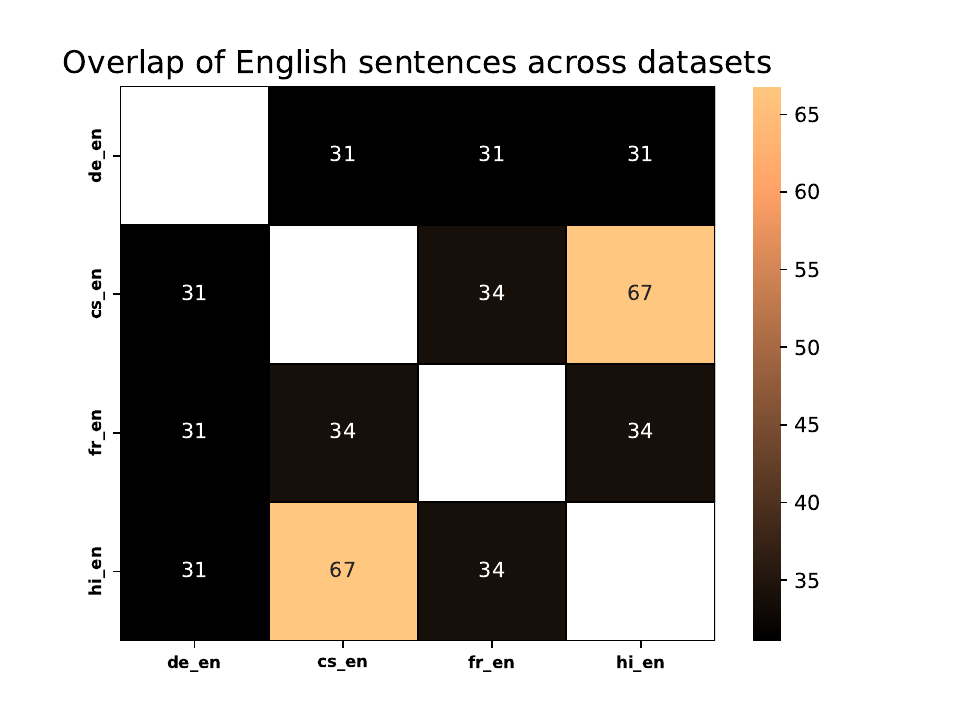}
    \caption{Overlap of English sentences}
    \label{fig:eng_sim}
\end{figure}

Next, we compare English sentences across different test sets. \Cref{fig:eng_sim} shows the extent of overlap between the sentences across the 4 different English test-sets used for the analysis. The hypothesis is that the analysis should show if there were dedicated layers for language specific shallow processing in the detectors or combinators. 

We see from \Cref{fig:det_sim_eng} that detectors in the early and middle layers of all models exhibit high degrees of rank correlation. We had already posited that the early detectors are multilingual. We use these results to extend that argument and state that this indicates the presence of ``shallow'' detectors i.e. they do not deal with the `semantic' content of the sentences. On the other hand, the combinators (\Cref{fig:com_sim_eng}) exhibit the maximal rank correlation in the later layers (close to the output). We posit that this indicates the presence of greater language specific neurons close to the output.   

\begin{figure}[ht]  
    \centering
    \begin{subfigure}[b]{0.2\textwidth}
        \centering
        \includegraphics[scale=0.24]{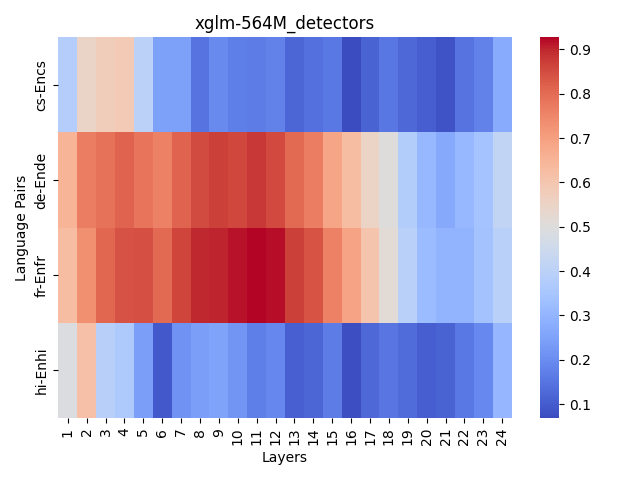}
    \end{subfigure}
    \hfill 
    \begin{subfigure}[b]{0.2\textwidth}
        \centering
        \includegraphics[scale=0.24]{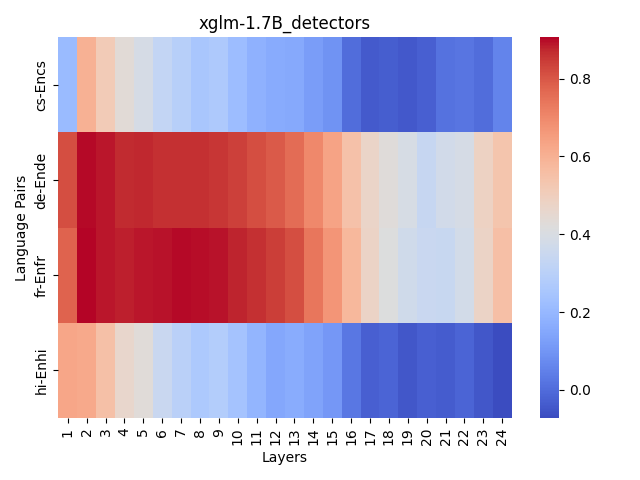}   
    \end{subfigure}
    \begin{subfigure}[b]{0.2\textwidth}
        \centering
        \includegraphics[scale=0.24]{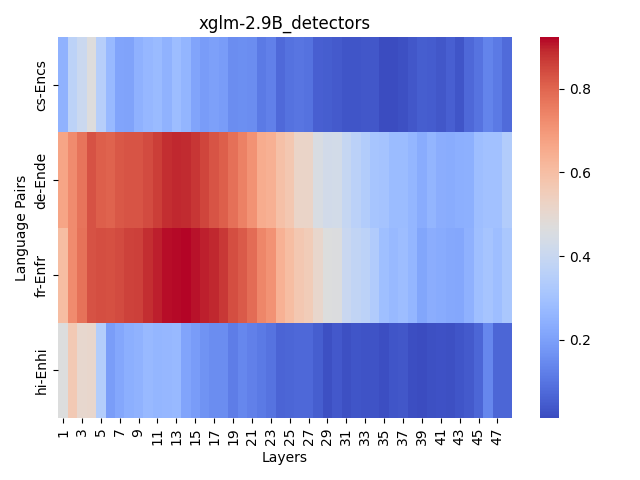}
    \end{subfigure}
    \hfill
    \begin{subfigure}[b]{0.2\textwidth}
        \centering
        \includegraphics[scale=0.24]{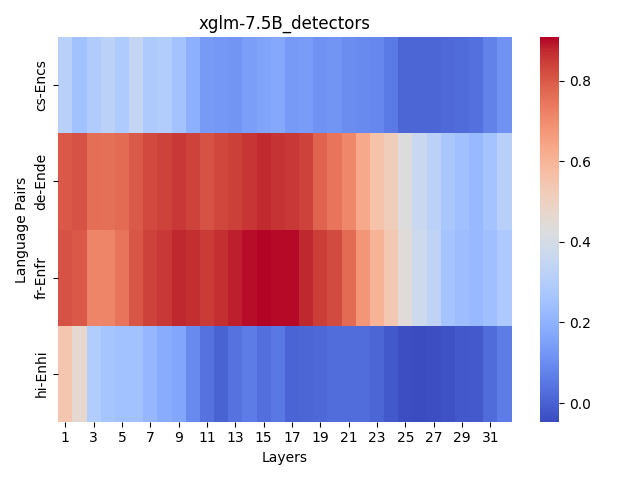}
    \end{subfigure}   
    \caption{Similarity of ranks (parallel data) : detectors}
    \label{fig:det_sim_par}
\end{figure}

\begin{figure}[ht]  
    \centering
    \begin{subfigure}[b]{0.2\textwidth}
        \centering
        \includegraphics[scale=0.24]{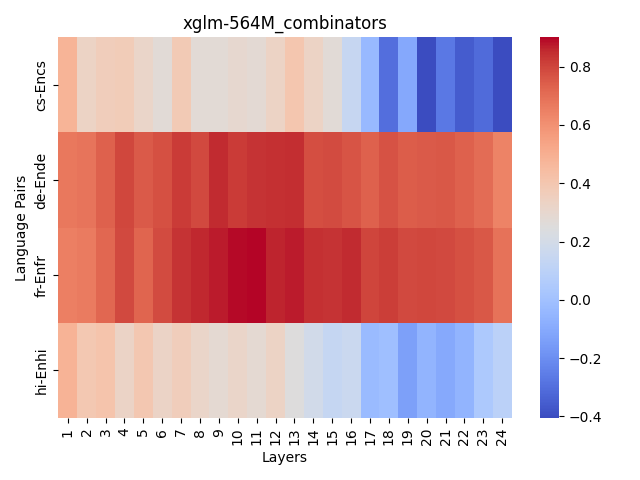}
    \end{subfigure}
    \hfill 
    \begin{subfigure}[b]{0.2\textwidth}
        \centering
        \includegraphics[scale=0.24]{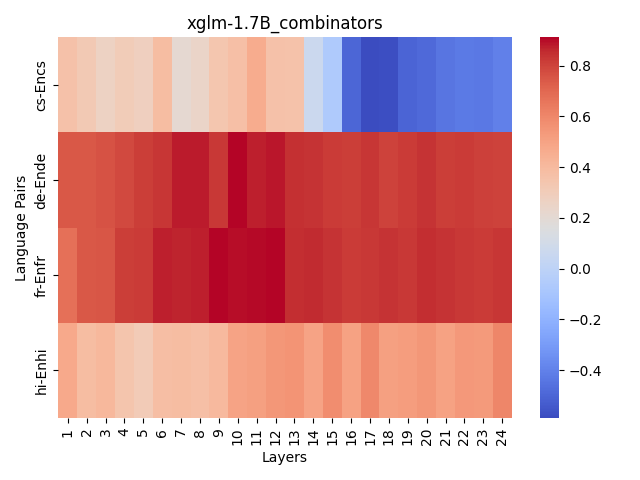}   
    \end{subfigure}
    \begin{subfigure}[b]{0.2\textwidth}
        \centering
        \includegraphics[scale=0.24]{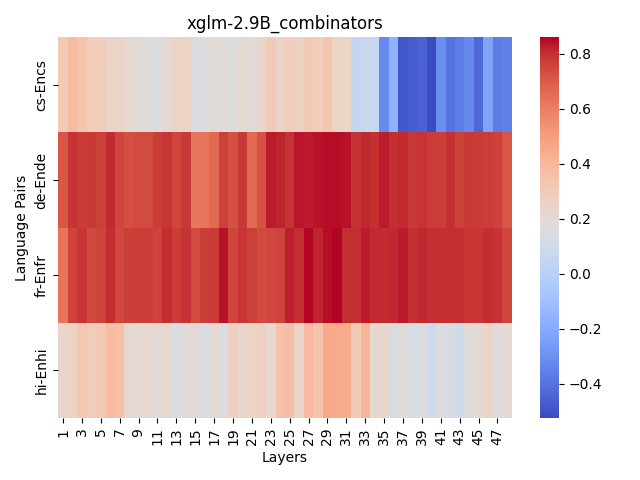}
    \end{subfigure}
    \hfill
    \begin{subfigure}[b]{0.2\textwidth}
        \centering
        \includegraphics[scale=0.24]{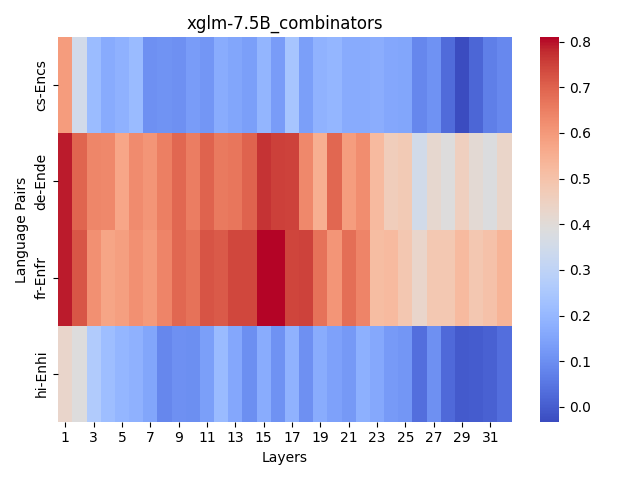}
    \end{subfigure}   
    \caption{Similarity of ranks (parallel data) : combinators}
    \label{fig:com_sim_par}
\end{figure}

\paragraph{Parallel Data Comparison}
Finally, we compare languages from parallel test-sets. The idea is to observe if certain layers processed ``semantic'' content similarly for certain models. We observe that for the English-German and English-French data, the detectors exhibit high degree of correlation in the early layers reaffirming earlier observations about their multilinugal and shallow processing abilities. For the combinators we find high values of rank correlation across layers for smaller models. The correlation is in fact highest when comparing German, French and English sentences. The combinators of the 7.5B parameter model however exhibits rank correlation close to zero across all layers for all language pairs except German-English and French-English. Infact, for 7.5B, the similarity is the most at layer 15, which also recorded the maximal representational similarity in \Cref{fig:com_dist}. We posit that like detectors, the combinators are multilinugal at early layers. The fact that the last layers of the XGLM models are language specific is pretty interesting as it has already been observed that for BERT, the upper layers (close to the ouptut) are more language specific. We note that this is an interesting parallel between the two kinds of LLMs (i.e. autoregressive versus auto-encoder).

\section{Conclusion}
In this work, we present our observations from analysing the FFN layers of various models from the XGLM language family. We first identify that across models, FFN representations are more sparse near the input and the output layers. Adopting the detectors-combinators view of FFNs and looking deeper into the nature of distribution of the activations, we find that for detectors, the activations get peaked through the layers. For the combinators, we observe that the activations get peaked across languages around layer 20, irrespective of the model depth, model dimension size or the size of the hidden representation in the FFNs. For combinators in fact, we discover that the representations across languages are very similar until very deep layers. Finally, we find that the detectors in the early layers process shallow features and are multilingual. They become language specific close to the output layers. We also find that the combinators are mostly multi-lingual around the middle layers.

\section{Limitations}
While our study offers insights into the multilingual capabilities of the XGLM model, several limitations must be acknowledged. 

Firstly, our analysis relies only snapshots of the model's activations, which does not fully capture the complexities of LLMs. In addition to model snapshots, probing tasks or even looking at the attention matrices could provide a more comprehensive understanding of how the model processes different languages.

Secondly, our investigation is limited to the XGLM model, and our findings may not generalize to other LLM architectures or implementations. 

Furthermore, our analysis is limited to a set of predefined languages and test sets, which may not fully represent the diversity of languages and linguistic phenomena captured by these models. Extending our analysis to more languages and exploring cross-linguistic variation in more detail could yield further insights into the generalizability of our findings.